\documentclass[a4paper, 10pt, conference]{ieeeconf}      
\usepackage{FG2024}
\makeatletter
\let\NAT@parse\undefined
\makeatother
\usepackage[backref=page,colorlinks=true]{hyperref}
\usepackage{cite}
\usepackage{caption}

\FGfinalcopy 

\usepackage{graphicx} 
\usepackage{float}
\usepackage{amsmath} 
\usepackage{amssymb}  
\usepackage{booktabs}
\usepackage{xr}

\title{\LARGE \bf
Giving a Hand to Diffusion Models: a Two-Stage Approach to Improving Conditional Human Image Generation
}

\author{\parbox{16cm}{\centering
   {\large Anton Pelykh, Ozge Mercanoglu Sincan, Richard Bowden}\\
   {\normalsize
   University of Surrey, United Kingdom}\\
   {\normalsize \{a.pelykh, o.mercanoglusincan, r.bowden\}@surrey.ac.uk}\\
   }
}

\begin{document}

\pagestyle{plain} 

\newcommand\myworries[1]{\textcolor{red}{#1}}
\newcommand\makepoint[1]{\textcolor{brown}{#1}}
\newcommand\highlight[1]{\textcolor{blue}{#1}}

\twocolumn[{%
\renewcommand\twocolumn[1][]{#1}%
\vspace{-10pt}
\maketitle
\thispagestyle{plain}
\vspace{-20pt}
\begin{center}
    \centering
    \captionsetup{type=figure}
    \setlength{\tabcolsep}{0pt} 
    \begin{tabular}{ccccccc}
        \includegraphics[width=0.14\linewidth, height=0.14\linewidth]{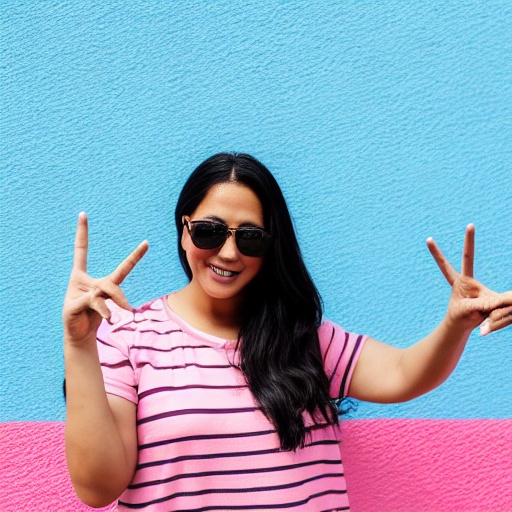} &
        \includegraphics[width=0.14\linewidth, height=0.14\linewidth]{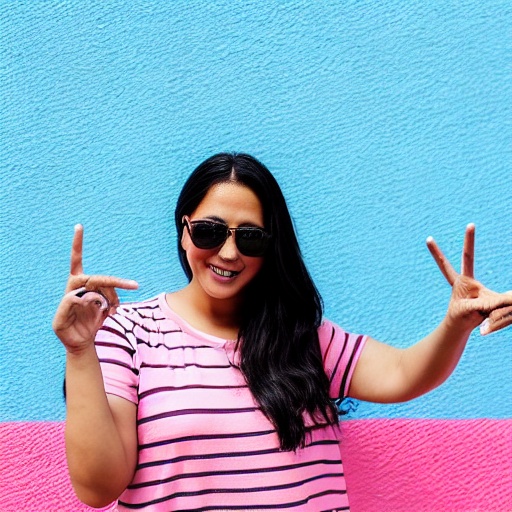} &
        \includegraphics[width=0.14\linewidth, height=0.14\linewidth]{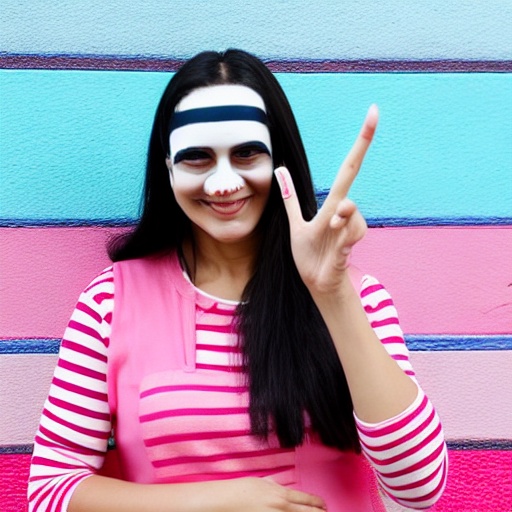} &
        \includegraphics[width=0.14\linewidth, height=0.14\linewidth]{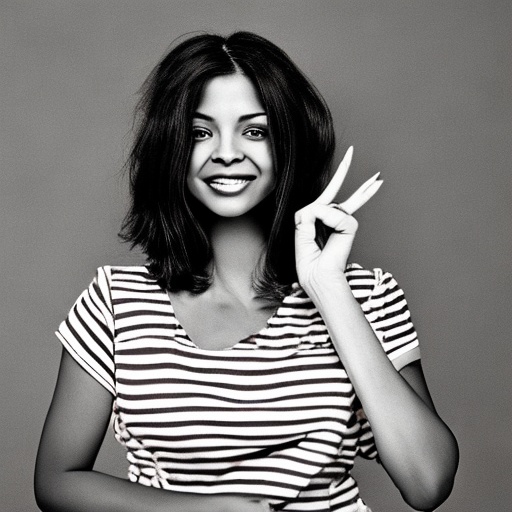} &
        \includegraphics[width=0.14\linewidth, height=0.14\linewidth]{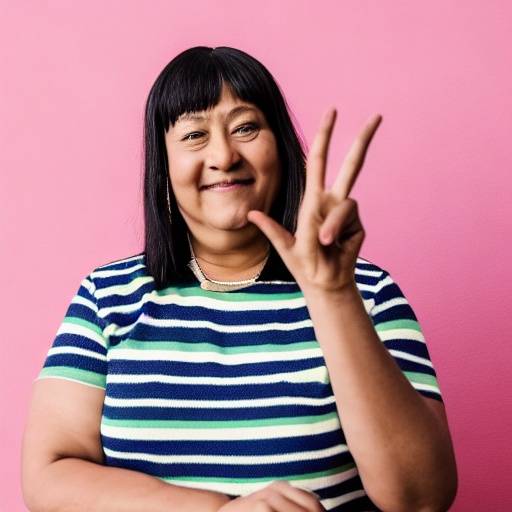} &
        \includegraphics[width=0.14\linewidth, height=0.14\linewidth]{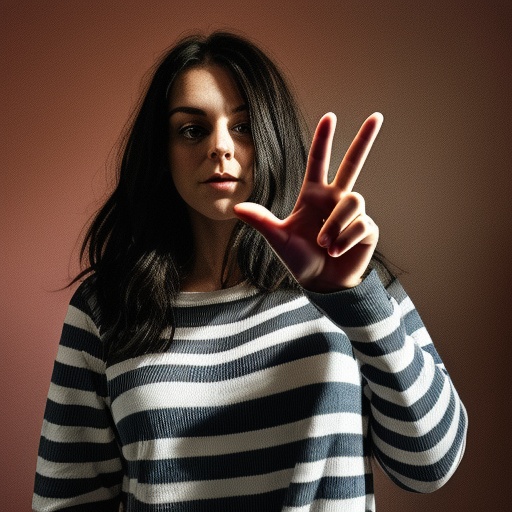} &
        \includegraphics[width=0.14\linewidth, height=0.14\linewidth]{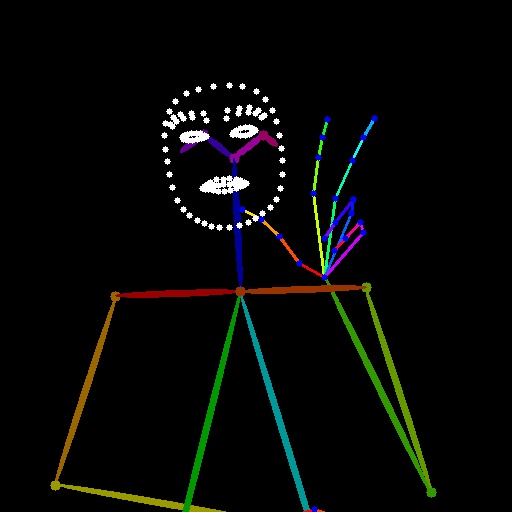}  \\

        \includegraphics[width=0.14\linewidth, height=0.14\linewidth]{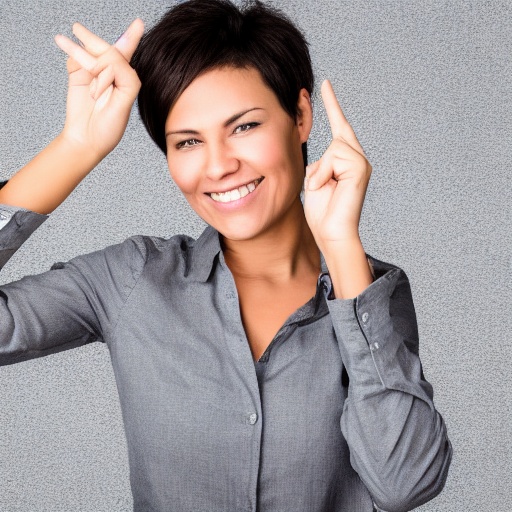} &
        \includegraphics[width=0.14\linewidth, height=0.14\linewidth]{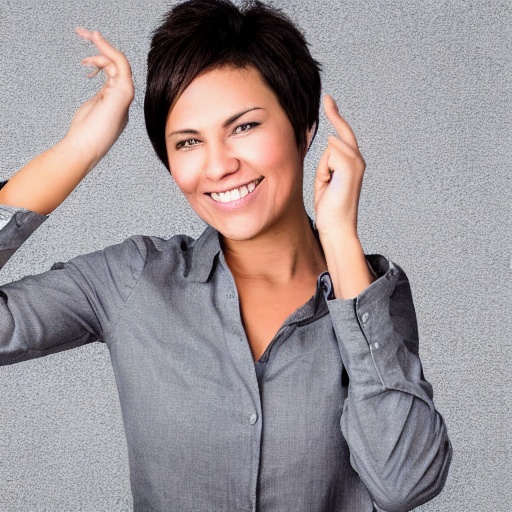} &
        \includegraphics[width=0.14\linewidth, height=0.14\linewidth]{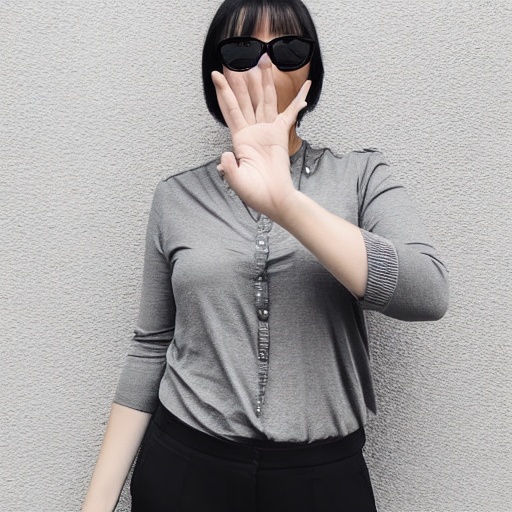} &
        \includegraphics[width=0.14\linewidth, height=0.14\linewidth]{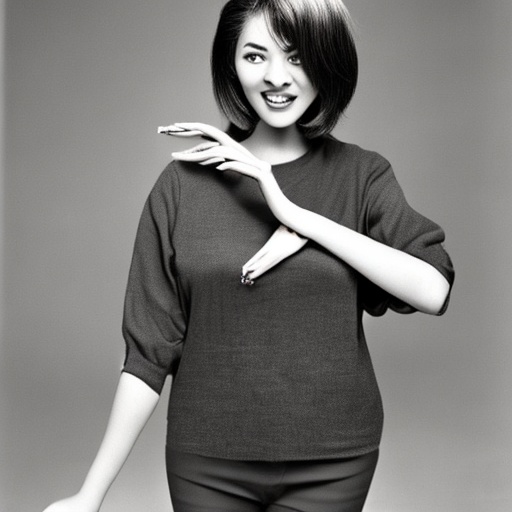} &
        \includegraphics[width=0.14\linewidth, height=0.14\linewidth]{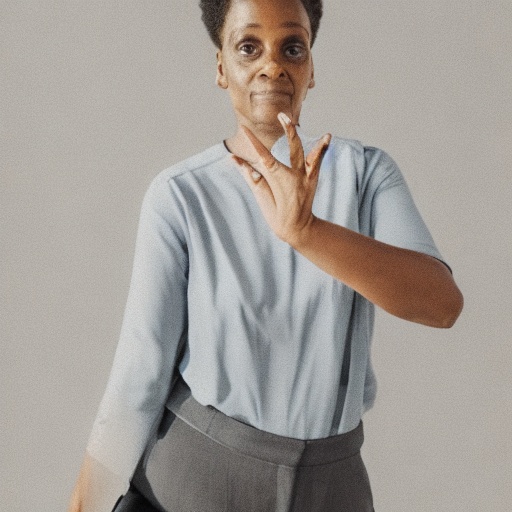} &
        \includegraphics[width=0.14\linewidth, height=0.14\linewidth]{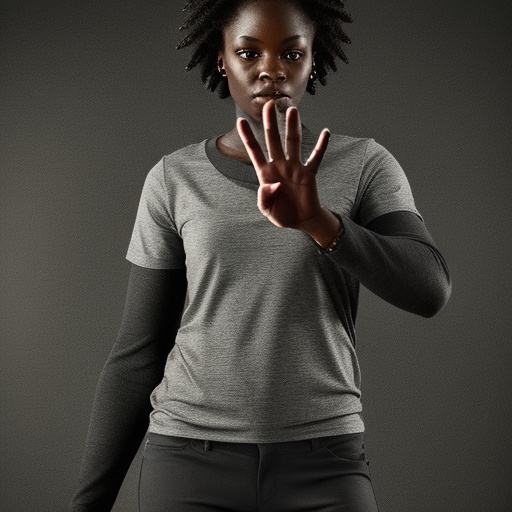} &
        \includegraphics[width=0.14\linewidth, height=0.14\linewidth]{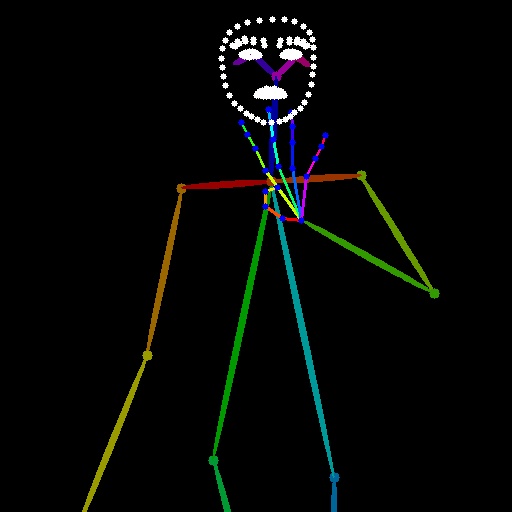} \\

        \includegraphics[width=0.14\linewidth, height=0.14\linewidth]{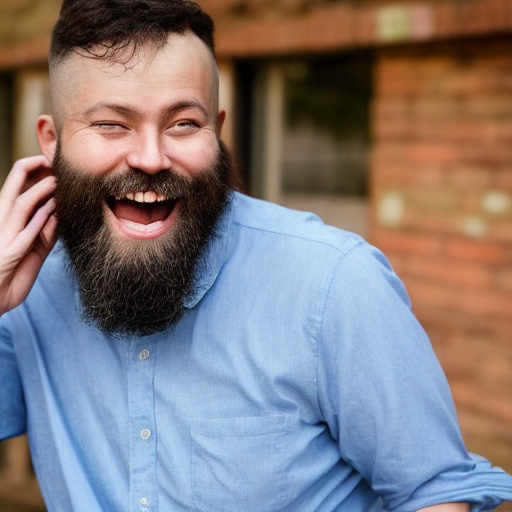} &
        \includegraphics[width=0.14\linewidth, height=0.14\linewidth]{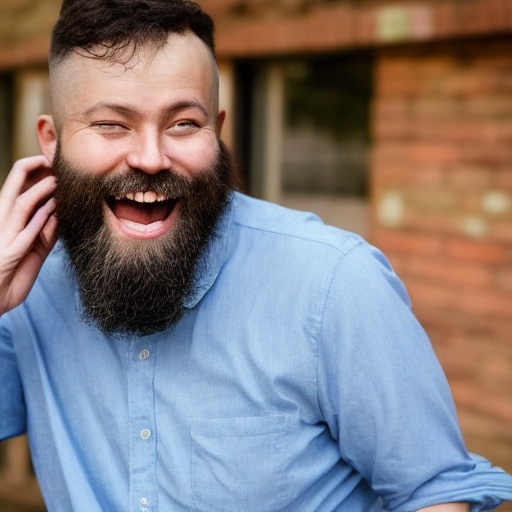} &
        \includegraphics[width=0.14\linewidth, height=0.14\linewidth]{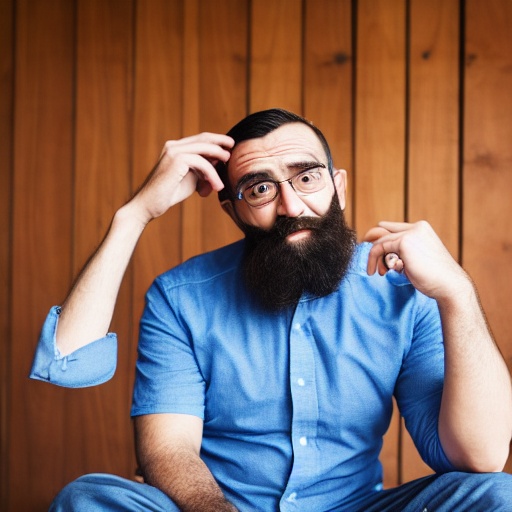} &
        \includegraphics[width=0.14\linewidth, height=0.14\linewidth]{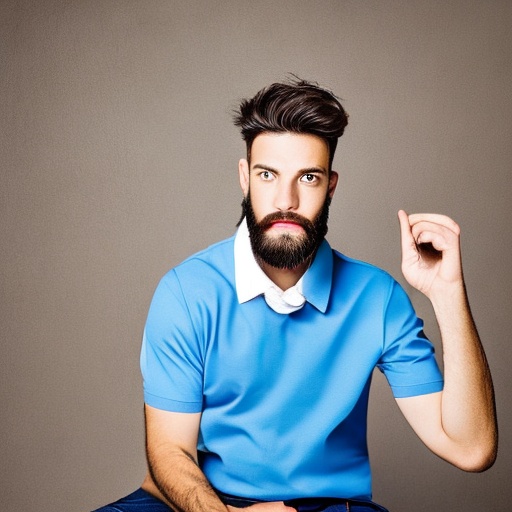} &
        \includegraphics[width=0.14\linewidth, height=0.14\linewidth]{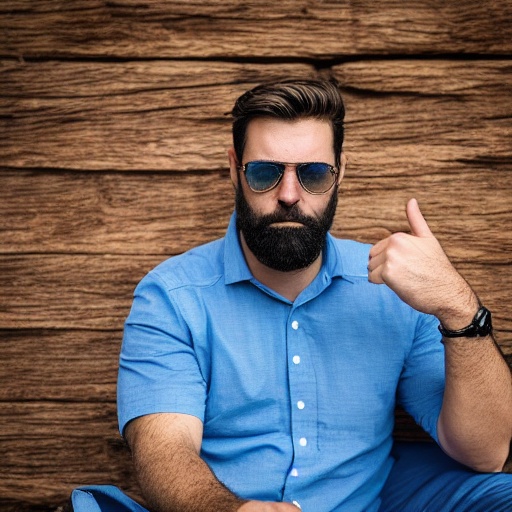} &
        \includegraphics[width=0.14\linewidth, height=0.14\linewidth]{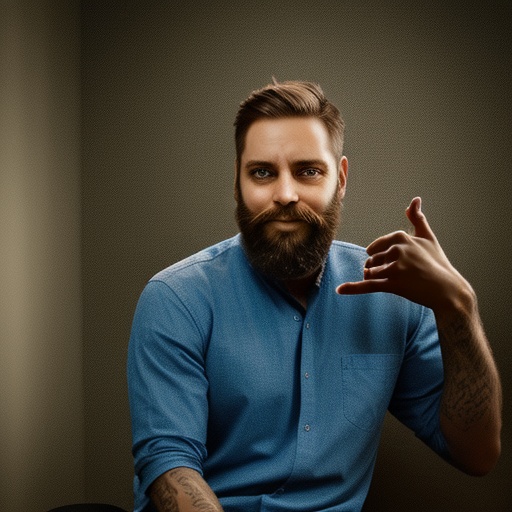} &
        \includegraphics[width=0.14\linewidth, height=0.14\linewidth]{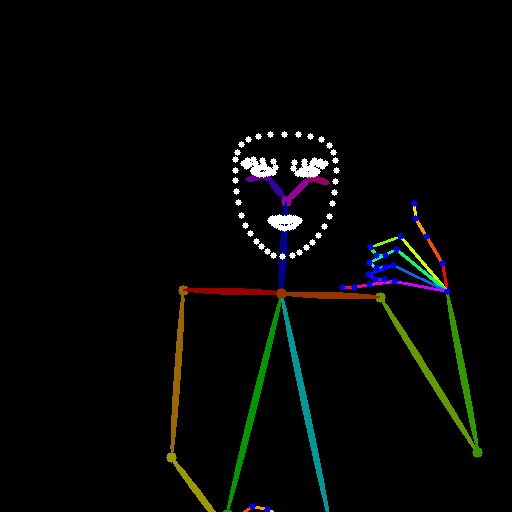} \\
        
        {\footnotesize Stable Diffusion \cite{rombach2022high}}  & {\small HandRefiner \cite{lu2023handrefiner}}  & {\small T2I-Adapter \cite{mou2024t2i}}  & {\small HumanSD \cite{ju2023humansd}} & {\small ControlNet \cite{zhang2023adding}} & {\small \textit{Ours}} & {\small Pose condition} \\
    \end{tabular}
    \captionof{figure}{Examples of images, generated by the proposed method (column 6) and the state-of-the-art diffusion models (columns 1 to 5), given the pose condition (final column) and the text description. The text prompts are provided in the supplementary material.}
    \label{fig:teaser}
\end{center}%
}]

\begin{abstract}

Recent years have seen significant progress in human image generation, particularly with the advancements in diffusion models. However, existing diffusion methods encounter challenges when producing consistent hand anatomy and the generated images often lack precise control over the hand pose. To address this limitation, we introduce a novel approach to pose-conditioned human image generation, dividing the process into two stages: hand generation and subsequent body outpainting around the hands. We propose training the hand generator in a multi-task setting to produce both hand images and their corresponding segmentation masks, and employ the trained model in the first stage of generation. An adapted ControlNet model is then used in the second stage to outpaint the body around the generated hands, producing the final result. A novel blending technique is introduced to preserve the hand details during the second stage that combines the results of both stages in a coherent way. This involves sequential expansion of the outpainted region while fusing the latent representations, to ensure a seamless and cohesive synthesis of the final image.
Experimental evaluations demonstrate the superiority of our proposed method over state-of-the-art techniques, in both pose accuracy and image quality, as validated on the HaGRID dataset. Our approach not only enhances the quality of the generated hands but also offers improved control over hand pose, advancing the capabilities of pose-conditioned human image generation. The source code is available here. \footnote{\url{https://github.com/apelykh/hand-to-diffusion}}

\end{abstract}

\section{INTRODUCTION}

Controllable human image generation is an important task in the field of visual content production with applications in advertising, game character creation and E-commerce amongst others. In recent years, diffusion models have overtaken the field with their flexibility and unprecedented quality of results. They dominate over other generative model types such as generative adversarial networks (GAN) and variational auto-encoders (VAE) \cite{dhariwal2021diffusion}. Many works have also explored ways to add pose control to diffusion generators \cite{zhang2023adding, mou2024t2i, ju2023humansd, baldrati2023multimodal, shen2023advancing}. Some approaches \cite{zhang2023adding, mou2024t2i} add a trainable branch on top of a frozen pre-trained Stable Diffusion (SD) model \cite{rombach2022high}. Other works \cite{ju2023humansd, baldrati2023multimodal} propose natively guided diffusion models that receive conditioning in concatenation with the input. 

Despite the remarkable results and increased flexibility of pose-guided approaches, diffusion models often struggle to achieve high-quality hand generation, leading to inaccuracies such as extra or missing fingers, distorted hand poses, and the presence of visual artifacts (see Fig. \ref{fig:teaser}). As the human brain is highly attuned to recognizing the details of human anatomy, including the structure and shape of hands, such failure cases are easily spotted and perceived as unnatural and eerie. In addition, modern diffusion generators do not provide precise control over hand pose and struggle to model hand interactions due to possible occlusions and the anatomic complexity of hands \cite{ju2023humansd, lu2023handrefiner}.

The fact that hands are high-fidelity objects with many degrees of freedom and a high probability of occlusion makes it a challenge to generate them realistically. At the same time, maintaining the model's generalization in terms of appearance and visual style while producing consistent hand anatomy is challenging. Training datasets rarely combine the volume and diversity of samples needed with high curation quality and precise annotation. Publicly available datasets that include annotated hands and hand interactions often lack visual diversity and tend to concentrate only on hands without including the rest of the body \cite{moon2020interhand2, moon2023reinterhand, Freihand2019}. Using such narrow-domain datasets, that do not present high variety in appearance and style for fine-tuning a pre-trained diffusion model, is typically detrimental. It can lead to decreased generality and expressiveness of the generator. This effect is referred to as "catastrophic forgetting" in the literature \cite{ju2023humansd}.

Recent works such as HandRefiner \cite{lu2023handrefiner} and Concept Sliders \cite{gandikota2023sliders} attempt to fix the quality of hand generation in SD and SDXL \cite{podell2023sdxl} respectively. Gandikota \textit{et al.} identify a low-rank direction in the parameter space of a diffusion model that targets hand quality and allow it to be modified via a Concept Slider. On the other hand, in HandRefiner Lu \textit{et al.} propose to repaint hands in the generated images with a depthmap-conditioned ControlNet \cite{zhang2023adding}. While both approaches show improved hand quality, they aim for general visual plausibility and do not allow pose control, which is a paramount factor in numerous application areas of generative models.

This work tackles the problem of high-quality hand generation in diffusion models while aiming to deliver precise control over the pose and preserve generality and high visual controllability. To our best knowledge, it is the first diffusion approach capable of high quality hand generation with pose control. This is achieved by dividing the task into two sub-tasks, namely hand generation and body outpainting around the generated hands. Such an architectural decision is motivated by the idea of decreasing the variability of data that the hand generator needs to learn from and allowing it to favor pose precision and articulation. At the same time, the outpainting stage leverages a conditional diffusion model that is tuned to accommodate complex hand shapes and is capable of synthesizing diverse appearances and styles. The hand generator is trained in a multi-task setting to produce a segmentation mask along with the main denoising objective, enabling precise body outpainting. To bring the two sub-components together in a coherent way and mitigate artifacts on the mask border, we propose a blending approach that utilizes sequential mask expansion.

The contributions of this work are summarized as follows:
\begin{itemize}
    \item We propose a novel two-stage diffusion-based approach to human image generation that is capable of producing high-quality hands with precise control over their pose.
    \item We show that conditional diffusion models can be successfully trained in a multi-task setting, predicting both the added noise and the semantic segmentation mask of the generated object.
    \item We introduce a blending technique that relies on sequential expansion of the outpainting region. It enables harmonious fusion of the both stages of the generation process while ensuring seamless transition between regions and preservation of detail.
    \item To demonstrate the effectiveness of the proposed solution, we conduct extensive experiments and comparisons with state-of-the-art models measuring pose precision, including a separate evaluation of hand pose, image quality and text-image consistency.
\end{itemize}

\section{RELATED WORK}
\label{sec:related_work}

\subsection{Image Generation with Diffusion Models}

Recently, there has been a significant increase of interest in diffusion models from the computer vision community due to their flexibility and high quality of results, that often dominate other generative model types \cite{dhariwal2021diffusion}. A notable branch of research in diffusion models is Denoising Diffusion Probabilistic Models (DDPM) \cite{sohl-dickstein2015-noneq-thermodyn, ho2020denoising} that utilize two Markov chains: a forward chain that noises the data, and a reverse chain that recovers data from noise.
Ho \textit{et al.} \cite{ho2020denoising} and Dhariwal and Nichol \cite{dhariwal2021diffusion} demonstrated the capability of denoising diffusion models to generate high-quality samples unconditionally with Song \textit{et al.} \cite{song2020denoising} and Nichol and Dhariwal  \cite{nichol2021-improved-ddpm} further proposing optimizations to the inference that allow for a significant speed up of the generation process. GLIDE \cite{Nichol2021-GLIDE} combined a diffusion model with text conditioning by encoding the input prompt into a sequence of embeddings with a transformer, which is then concatenated with the attention context of each layer. Similarly, DALL-E2 \cite{ramesh2022hierarchical} and Imagen \cite{saharia2022photorealistic} employed a modified GLIDE architecture to map the the CLIP \cite{radford2021learning} and T5-XXL encoder \cite{Raffel2020-T5} embedding space correspondingly into the image space via the reverse diffusion process, generating images that convey the semantic information of the input caption. While early diffusion approaches were performed in pixel space, Rombach \textit{et al.} \cite{rombach2022high} proposed Latent Diffusion by moving the denoising process to the lower-dimensional latent space of a pre-trained autoencoder, benefiting from perceptual compression of the modeled data and unlocking greater flexibility for solving various image-to-image and text-to-image tasks.

\subsection{Pose-Conditioned Human Image Generation} 

Although unconditional and text-conditioned approaches can often produce high-quality realistic results, the limited control over generation makes such models unusable for many content production use-cases.

Generative Adversarial Networks (GAN) \cite{goodfellow2014generative} have been widely used to introduce pose control to image generation. Ma \textit{et al.} \cite{ma2017pose-guided-person} utilized explicit appearance and pose conditioning of a two-stage GAN architecture to generate difference maps between the coarsely generated image and the target to ensure faster model convergence. In contrast, Siarohin \textit{et al.} \cite{siarohin2018deformable-gans} proposed an end-to-end method that explicitly models pose-related spatial deformations by employing deformable skip-connections. However, their method requires extensive computations of affine transformations to solve pixel-level misalignment caused by pose differences. Zhu \textit{et al.} \cite{zhu2019progressive-pose-attention} introduced a progressive scheme that transfers the initial pose to the target through a sequence of intermediate representations using \textit{Pose-Attentional Transfer Blocks}. Subsequently, Zhou \textit{et al.} \cite{zhou2022cross} proposed a cross-attention-based module that distributes features from semantic regions of the source image to satisfy the target pose instead of directly warping the source features. The aforementioned approaches were mostly developed using fashion datasets and/or low-resolution images without account for the hand pose. With this in mind, Saunders \textit{et al.} proposed GAN-based methods \cite{saunders2022signing, saunders2020everybody} for sign language applications that aim to generate fine-grained hand details. Although the approaches explicitly model hands, they can only produce appearances seen during training and do not generalize to out-of-distribution visual conditions.


Diffusion models have seen extensive use for pose-conditioned human image generation. Bhunia \textit{et al.} \cite{bhunia2023person} achieve pose control by concatenating the skeleton condition to the model input. Additionally, the style image features are passed to cross-attention blocks to better exploit the correspondences between the source and target appearances. Building up on Latent Diffusion \cite{rombach2022high}, a number of works \cite{zhang2023adding, mou2024t2i, ju2023humansd, baldrati2023multimodal} extended it to condition the denoising process on various modalities such as human pose keypoints, sketches, edge maps, depth maps, color palette etc. ControlNet \cite{zhang2023adding} introduces a trainable copy of a Stable Diffusion (SD) encoder to extract features from the condition while keeping the base model frozen during training. Similarly, T2I-Adapter \cite{mou2024t2i} uses lightweight composable adapter blocks for condition feature extraction which can be combined for multi-condition setting. In \cite{zhang2023adding} and \cite{mou2024t2i}, the features learned by encoders are combined with the features of the frozen backbone model in an additive manner which may provoke trainable-frozen branch conflicts, as discussed in HumanSD \cite{ju2023humansd}. To mitigate this issue, Ju \textit{et al.} do not employ additional encoders and make all the parameters of the underlying Stable Diffusion model trainable, while trying to mitigate the issue of catastrophic forgetting by using the proposed heatmap-guided denoising loss. They achieve pose control by concatenating the skeletal condition with noisy input latents. Similarly, Baldrati \textit{et al.} \cite{baldrati2023multimodal} extend the input of the SD model to include the human pose image and garment sketch, with textual description being encoded with CLIP and passed into the model through a cross-attention mechanism.
Notably, most of the recent pose-conditioned approaches to image generation \cite{bhunia2023person, mou2024t2i, ju2023humansd, baldrati2023multimodal, shen2023advancing} do not include hand keypoints into a skeleton representation and therefore do not offer control over the hand pose. On the other hand, the models that offer such control, e.g. ControlNet, fail to produce realistic and anatomically correct hands. This shortcoming is tackled by our proposed approach.




\section{PROPOSED METHOD}
\label{sec:methodology}

\begin{figure*}
    \begin{center}
        \centering
        \includegraphics[width=0.9\textwidth]{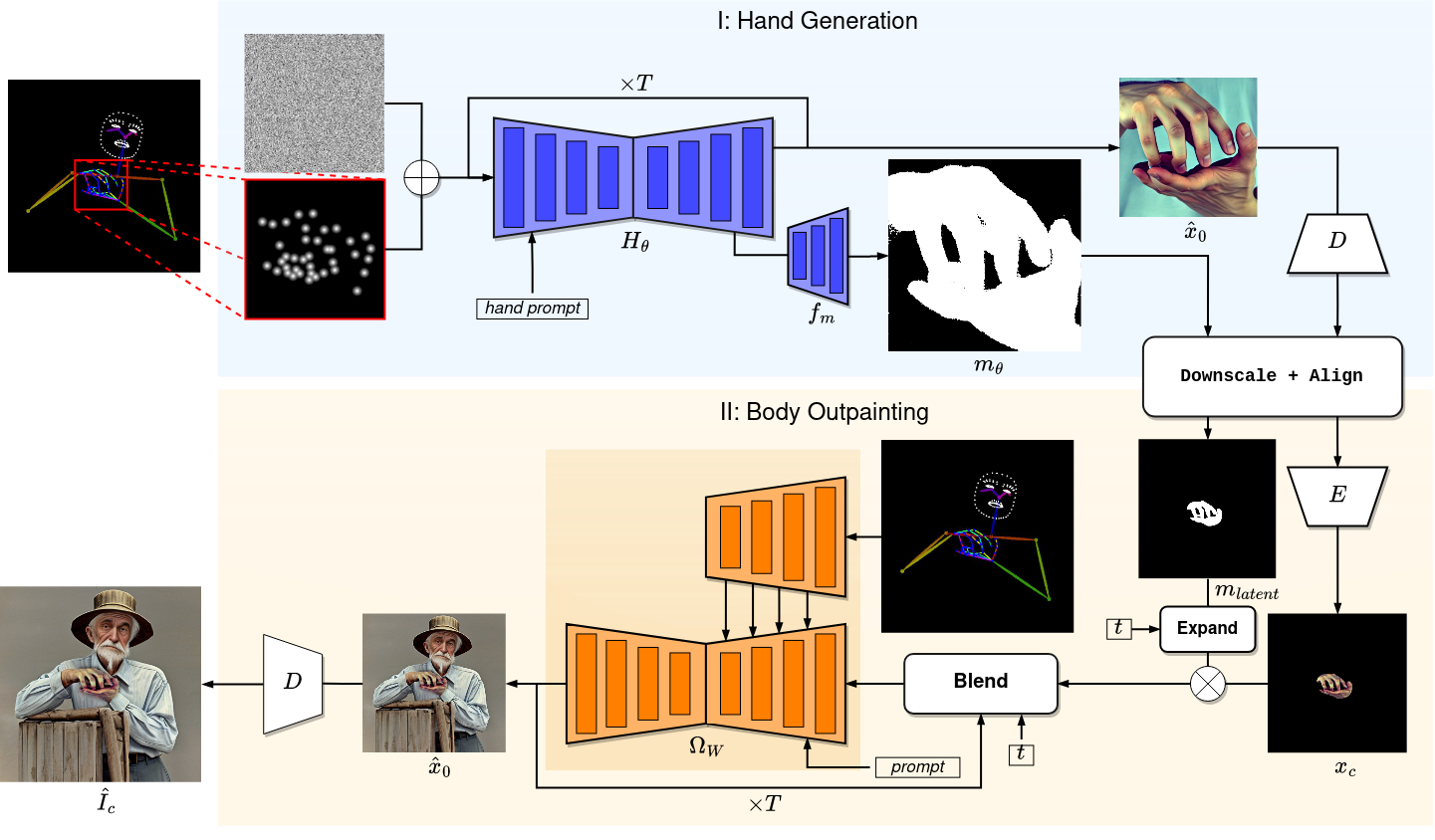}
        \caption{General overview of the proposed approach. We divide image generation into two sub-tasks: (I) hand generation (top part) and (II) body outpainting around the hands (bottom part).}
        \label{fig:system-overview}
    \end{center}%
\end{figure*}

A general overview of the proposed framework is shown in Fig. \ref{fig:system-overview}. In this work we propose to break the image generation task into two sub-problems: hand generation and body outpainting around the hands. Firstly, the hand image and the corresponding segmentation mask are produced by the diffusion-based generator, guided with the hand pose condition in the form of a keypoint heatmap. It is possible due to the multi-task training setting that we propose for the generator. The obtained results are further resized and aligned with the global body skeleton to serve as input to the outpainting module. The final image is produced during the second stage by outpainting around the generated hands using the ControlNet \cite{zhang2023adding} model. It is guided by the skeleton image and the segmentation mask, obtained at the previous stage. We are able to bring the two stages together in a harmonious way by using the proposed blending strategy with sequential mask expansion. The division into two sub-components intends to decrease the complexity of the task for the hand generator, enabling it to prioritize pose precision and articulation. At the same time, employing a separate model for the outpainting stage allows our system to have greater control over the generated appearance and style via the text prompt.

In Section \ref{sec:latent_diffusion} we outline the Latent Diffusion paradigm that serves as the foundation for our work. Sections \ref{sec:hand_generation} and \ref{sec:body_outpainting} further describe each stage of the generation process in detail, while Section \ref{sec:blending} explains the proposed blending technique that unifies both sub-components and helps to produce a coherent output image.

\subsection{Latent Diffusion Models}
\label{sec:latent_diffusion}

The idea of Latent Diffusion \cite{rombach2022high} is to perform the diffusion process in the latent space of a pre-trained autoencoder to decrease the dimensionality of the data and operate on the feature level instead of raw pixels. The input image $I \in \mathbb{R}^{H \times W \times 3}$ is put through an encoder $E$ to obtain its latent representation $x_0 = E(I)$, where $x_0 \in \mathbb{R}^{\frac{H}{8} \times \frac{W}{8} \times 4}$. The latents are subsequently corrupted by noise, following the forward diffusion process, a Markov chain of Gaussian transitions: 
\begin{equation}
\label{eq:diffusion_forward_naive}
    q(x_{t}|x_{t-1}) =\mathcal{N}(x_{t};\sqrt{1-\beta_{t}}x_{t-1},\beta_{t}\textbf{I}),
\end{equation}
where $t = 1, ..., T$ is the time step that defines the strength of the added noise, $\beta_{t}$ is the noise variance and $q(x_{t}|x_{t-1})$ is the conditional probability of $x_t$ given $x_{t-1}$. By utilizing the properties of the above process and performing a reparameterization trick, we can obtain $x_t$ from any time step in the closed form:
\begin{equation}
\label{eq:diffusion_forward_closed_form}
    q(x_{t}|x_0)=\mathcal{N}(x_{t};\sqrt{\bar{\alpha_{t}}}x_0,(1 - \bar{\alpha_{t}})\textbf{I}),
\end{equation}
where $ \bar{\alpha_{t}} = \prod_{i=1}^{t}(1 - \beta_{i})$.

Our goal is to restore a clean sample $\hat{x_0}$ from the noise $x_T$. However, the reverse process $q(x_{t-1}|x_{t})$ is intractable in general case, therefore it is approximated with a Gaussian generative process that utilizes a U-Net model $\epsilon_{\theta}$, trained to predict the added noise and subsequently recover $\hat{x_0}$:
\begin{align}
    \label{eq:ddpm_backward}
    p_{\theta}(x_{0:T}) &= p(x_T) \prod_{t=1}^{T} p_{\theta}(x_{t-1}|x_t), \\
    p_{\theta}(x_{t-1}|x_t) &= \mathcal{N}(x_{t-1}; \mu_{\theta}(x_t, t), \Sigma_{\theta}(x_t, t)).
\end{align}
In DDPM \cite{ho2020denoising} the mean of the reverse diffusion process $\mu_{\theta}(x_t, t)$ is reparameterized in the following way:
\begin{equation}
    \mu_{\theta}(x_t, t) = \frac{1}{\sqrt{\alpha_t}} (x_t - \frac{1 - \alpha_t}{\sqrt{1 - \bar\alpha_t}}\epsilon_{\theta}(x_t, t)),
\end{equation}
where $\epsilon_{\theta}(x_t, t)$ is the predicted noise. DDIM \cite{song2020denoising} further generalizes DDPM and defines a family of non-Markovian processes with the sampling of a denoised element defined as:
\begin{equation}
\label{eq:ddim_backward}
\begin{split}
    p_{\theta}(x_{t-1}| x_t, x_0) &= \mathcal{N}(\sqrt{\alpha_{t-1}}\left( \frac{x_{t} - \sqrt{1 - \bar{\alpha}_t} \cdot \epsilon_{\theta}(x_t, t)}{\sqrt{\bar{\alpha}_t}} \right) \\
    &+ \sqrt{1 - \alpha_{t-1} - \sigma_{t}^2} \cdot \epsilon_{\theta}(x_t, t), \sigma_{t}^2\textbf{I}).
\end{split}
\end{equation}
DDIM significantly accelerates the generative process by considering sampling trajectories of length smaller than $T$ without retraining the model.

Latents $\hat{x_0}$, obtained from the denoising process, are then translated back to the pixel space using the decoder $D$ to form the resulting generated image $\hat{I} = D(\hat{x_0})$.

\subsection{Multi-Task Hand Generation}
\label{sec:hand_generation}

We use a pre-trained SD model as the foundation for the proposed hand generator $H_{\theta}$ and finetune it in a multi-task setting to predict the noise together with the segmentation mask of the generated hands. Baldrati \textit{et al.} \cite{baldrati2023multimodal} and Ju \textit{et al.} \cite{ju2023humansd} demonstrated that the SD architecture can be successfully conditioned by concatenating additional inputs without employing a separate encoder. Taking inspiration from \cite{baldrati2023multimodal} and \cite{ju2023humansd}, our hand generator accepts an additional condition $c_p \in \mathbb{R}^{K \times H_{h} \times W_{h}}$ that is concatenated with the noisy input latents, guiding the generation process towards the specified hand shape. The noise term $\epsilon_{\theta}(x_t, t)$ in (\ref{eq:ddim_backward}) is denoted here as $\hat{\epsilon}_{\theta}$, which is predicted by the proposed conditional model along with the segmentation mask $m_{\theta}$:
\begin{equation}
\label{eq:generator_outputs}
    \hat{\epsilon}_{\theta}, \highlight{m_{\theta}} \leftarrow H_{\theta}(x_t, t, \highlight{c_p}).
\end{equation}
At each step of the diffusion process, a denoised hand latent $\hat{x}_{0}$ is obtained with DDIM sampling (\ref{eq:ddim_backward}) using the predicted noise $\hat{\epsilon}_{\theta}$.

In this work, a conditional input $c_p$ has $K = 11$ channels, where 10 channels are occupied by the hand keypoint heatmap and 1 channel represents the hand segmentation mask. Each channel of the heatmap contains the keypoints of an individual finger to provide better separability in situations when fingers overlap or are occluded. Furthermore, a segmentation mask of the hands is included in the input to bring additional spatial guidance to the model. During training, the mask is filled with zeros with the probability $p = 0.5$ to increase the robustness of the model and enable mask-free inference. Both the keypoint heatmap and the hand segmentation mask are downsized to the latent dimension with bilinear interpolation to provide explicit pose and layout control to the generator. To accommodate the increased number of input channels, we extend the first convolutional layer of the pre-trained SD architecture with randomly initialized weights and further train the network.

Hand segmentation masks are predicted by a stack of 4 transposed convolutional layers $f_m$ with kernels of the size $2 \times 2$ and stride 2. The outputs of each non-final layer are passed through the Sigmoid Linear Unit (SiLU \cite{hendrycks2016gaussian}) activation function. The mask prediction head is built on top of the last layer of the SD decoder and it produces outputs in the spatial resolution of the input image $I$. The predicted mask is further used to define the target region for the body outpainting module and to blend hands and body in a harmonious way.

The network is trained using the combined objective:
\begin{align}
\label{eq:hand_generator_loss}
    L &= L_{LDM} + \lambda \frac{1}{N} \sum_{i=0}^{N} ( M_i - m_{\theta} )^{2}, \\
    L_{LDM} &= \mathbb{E}_{\epsilon \sim \mathcal{N}(0, 1), t \sim [1, T]}\left[ \lVert \epsilon - \hat{\epsilon}_{\theta} \rVert_{2}^{2}\right],
\end{align}
where $M_i$ is the ground truth segmentation mask for the $i$-th sample, $\lambda$ is a hyperparameter that defines the weight of the segmentation loss, $\hat{\epsilon}_{\theta}$, $m_{\theta}$ are the outputs of the hand generator, as shown in (\ref{eq:generator_outputs}). Apart from the practical use of the predicted segmentation mask in the next stage, including an extra objective provides an additional regularization to the training process, thus making the generator more robust \cite{he2017maskRCNN, Long2015LearningMT, Kendall2017MultitaskLU}.

\subsection{Body Outpainting}
\label{sec:body_outpainting}

Given the generated hand image and its predicted segmentation mask, the image background is removed. Both the resulting foreground image and the mask are further downscaled and aligned with the full body skeleton to form the canvas for outpainting $I_c$ and its corresponding mask. $I_c$ is then encoded into a latent space using the encoder $x_c = E(I_c)$, and the mask $m_{\theta c}$ is downsized to match the spatial dimensions of the latent representation:
\begin{equation}
    m_{\theta c} \in \mathbb{R}^{H \times W} \rightarrow m_{latent} \in \mathbb{R}^{\frac{H}{8} \times \frac{W}{8}}.
\end{equation}

The final generated image is obtained by painting the body around the hand region with a ControlNet model $\Omega_{W}$. The objective of the model is to predict the unknown latent pixels $(1 - m_{latent}) \odot x_c$ of the input canvas while leaving the masked area $m_{latent} \odot x_c$ unchanged, guided by the body pose in the form of a skeleton image and a mask of the target region. Even though the ControlNet receives the mask as a condition, a diffusion process is performed over the full area of the latent and therefore corrupting the hand region. To preserve the hand details, the latents at each step are obtained by blending the input canvas and the denoised latents at the current step, similarly to \cite{Avrahami2021BlendedDF, Avrahami2023blendedLatent}:
\begin{equation}
\label{eq:latent_blending}
    x_t = m_{latent} \odot x_c + (1 - m_{latent}) \odot x_t.
\end{equation}

The pre-trained skeleton-conditioned ControlNet model can naturally solve the inpainting task by noising, and subsequently restoring, the masked region of the input. However, in the case of body outpainting, it tends to hallucinate objects and unnatural backgrounds around the hand region as the model learned to associate non-neutral hand shapes to holding objects during the generic training. Furthermore, a pre-trained model often tries to complete the hand outside the bounds of the mask thus making it anatomically incorrect. To mitigate these issues, we fine-tune ControlNet for body outpainting by providing an initial canvas with hands, segmented from the original image, and tasking the model to complete the image by predicting the outside region. We provide the skeleton image as a condition into the model's encoder and blend the noisy and hand latents as described in (\ref{eq:latent_blending}). The masked $L2$ reconstruction loss is used for the training.

\subsection{Sequential Mask Expansion}
\label{sec:blending}

When outpainting the body around the previously generated hands, it is crucial to ensure hand detail preservation as well as a seamless transition and natural connectivity between the two regions. While the naive blending strategy described in (\ref{eq:latent_blending}) enforces the hand region stays the same throughout the diffusion process, it often leads to anomalies around the region border in the case of non-uniform background. Although tuning the ControlNet for body outpainting helps to alleviate this issue, the model still tends to expand the hand outside the masked region, add extra fingers or introduce erroneous textures for complex hand shapes.

To address the irregularities around the mask border, we propose to gradually dilate the input hand mask for $T$ iterations, where $T$ is the number of diffusion steps, and then use the expanded masks as $m_{latent}$ in (\ref{eq:latent_blending}) starting from the largest and arriving at the original at step $T$. At the same time, the underlying denoising UNet of the body outpainter receives a precise hand mask at every iteration of the diffusion process. The intuition behind this process is that the possible distortions that the model may manifest around the hand region will be replaced by the latent pixels from the uniform background of the initial canvas. At the same time, the replaced region will be harmonized and blended with the rest of the latent during the next step of diffusion. Using smaller masks for each diffusion step allows washing out the hard border of the expanded region and avoids visible edge artifacts. The last two iterations of diffusion are performed on the full latent without masking to further unify the two regions in terms of transition smoothness, color distribution and shadows.

In Blended Latent Diffusion \cite{Avrahami2023blendedLatent} progressive mask shrinking was employed to enable text-guided image editing in a thin masked region. However, our mask expansion approach is solving a conceptually different task of harmonious blending of two regions of the latent representation with no constraints on the size of the inpainted region. In our case, the diffused region typically spans most of the image and we are forcing it to envelop the generated hands in a coherent and artifact-free way by progressively expanding the mask.

After the diffusion process in completed, the denoised latents are mapped back to the pixel space using the decoder D, i.e $\hat{I_c} = D(\hat{x_0})$. We then blend the resulting image with the input hand region using the initial mask $m_c$ following the naive strategy from (\ref{eq:latent_blending}). This allows us to reintroduce sharpness to the hands that might have been reduced during the unmasked diffusion steps with no detrimental effects on the blending consistency.

\addtolength{\textheight}{-1cm} 

\section{EXPERIMENTS AND RESULTS}

\subsection{Datasets}
\label{sec:datasets}

\def \ih {InterHand2.6M}
\def \reih {Re:InterHand}
\def \hagrid {HaGRID}
\def \laion {LAION-Human}

\begin{figure}[b!]
    \centering
    \includegraphics[width=0.48\textwidth]{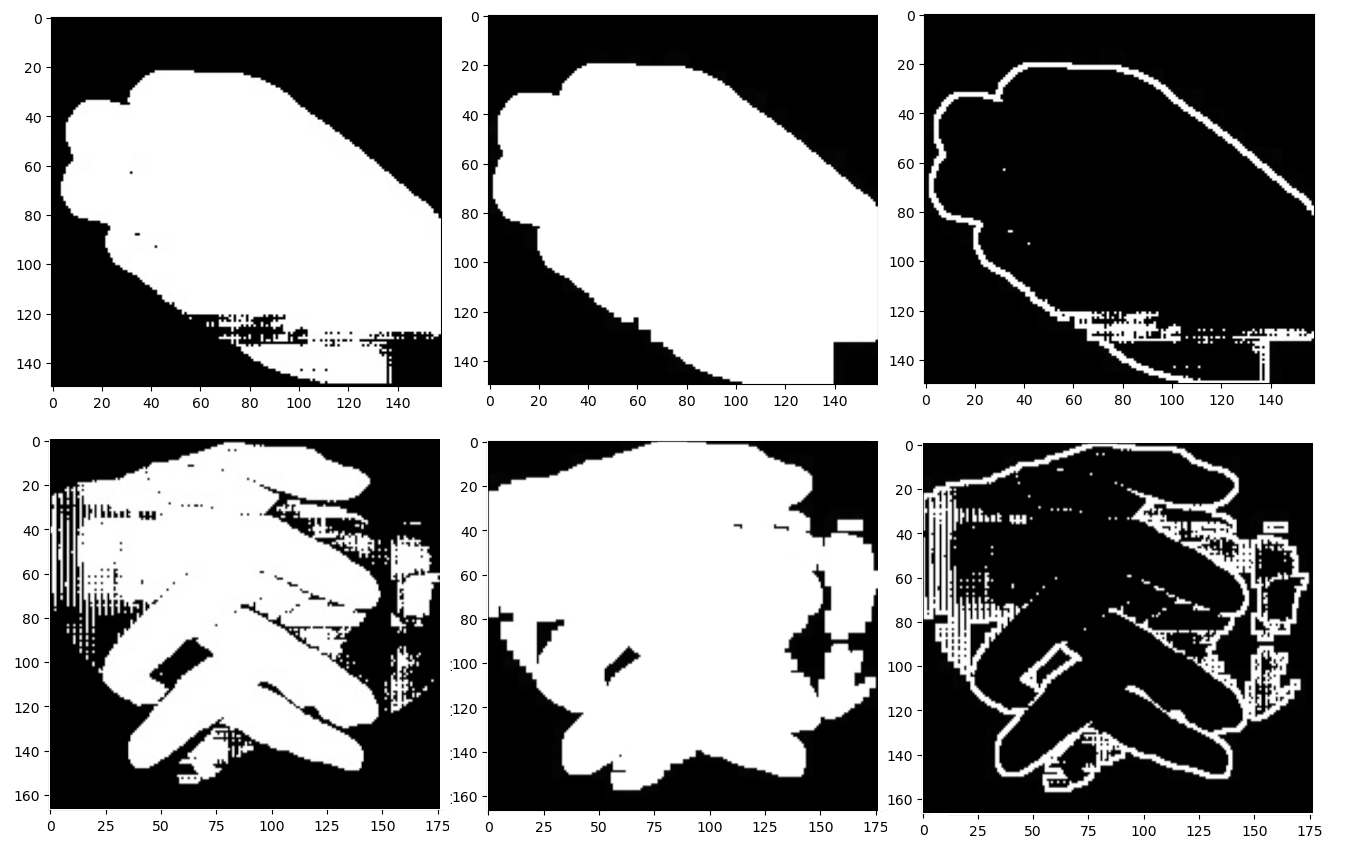}
    \caption{Segmentation masks extracted with SAM (left), masks after applying a dilation kernel (middle), pixel-wise difference between the two (right)}
    \label{fig:sam-masks-dilation}
\end{figure}

A combination of \ih{} \cite{moon2020interhand2}, \reih{} \cite{moon2023reinterhand} and \hagrid{} \cite{hagrid} datasets is used for training the hand generator. The datasets are combined to ensure overall sample quality and diversity. \ih{} is restricted to a studio environment with distinct lighting and a limited number of participants whereas \reih{} provides synthetic 3D renders of real images. \hagrid{} is the most diverse of the three datasets as it was captured ``in the wild'' but it includes images of varying quality and only bounding boxes as annotation. Both \ih{} and \reih{} provide precise hand keypoints and for \hagrid{} keypoints are extracted using the Mediapipe holistic model \cite{mediapipe}. The hand segmentation masks for \ih{} and \hagrid{} are obtained with SAM ViT-H \cite{kirillov2023segment} by using keypoints as queries for the model, while \reih{} includes the masks as a part of the dataset. The masks extracted with SAM often include checkerboard artifacts and discontinuities on the edges so they were processed with a $5\times5$ dilation kernel to mitigate this issue. Fig. \ref{fig:sam-masks-dilation} shows examples of masks with artifacts and their post-processed versions. We also use the LLaVA-v1.5-7b \cite{liu2024visual} model to produce image captions for \hagrid{}. Sequence-level captions for \ih{} and \reih{} are created manually to include gender, skin tone and the details of the hand appearance.

To train the hand generator, we crop the square hand regions from the original images and resized them to the resolution $512\times512$ to accommodate the pre-trained SD architecture. For cases where hands are interacting and their bounding boxes intersect, both hands are included in the same crop, otherwise one hand is cropped at random during training. We also apply RGB value shifting and random brightness and contrast changes to augment the training samples. The total training dataset size for the hand generator is $193,300$ samples, where $60,000$ are randomly sampled from \ih{}, $56,500$ from \reih{} and $76,800$ from \hagrid{} training subsets while keeping the original gesture distribution of \hagrid{}. 

To construct the dataset for training the oupainting model, we utilize \laion{} (from HumanSD \cite{ju2023humansd}). Similarly to how we process \hagrid{}, the keypoints are extracted with Mediapipe and the hand segmentation masks with SAM ViT-H. Although \laion{} includes text prompts and pose keypoints, the former are extremely noisy due to the automatic way in which the original dataset was constructed, and the latter are too sparse. Therefore, we replace the original text prompts with ones obtained from LLaVA-v1.5-7b, and use Mediapipe estimates as the ground-truth keypoints. The images are filtered to keep only those with a single human present in the frame. We further discard images for which SAM did not produce a hand segmentation mask of at least 2500 pixels. In this way we extract a total of $36,500$ images that are used for model training.

\subsection{Implementation Details}

The hand generator is initialized from the official SD v1.5 checkpoint and further tuned for 5 epochs ($30,000$ iterations) on the combination of \ih{}, \reih{} and \hagrid{}, as described in Section \ref{sec:datasets}. The segmentation mask loss weight $\lambda$ from (\ref{eq:hand_generator_loss}) is set to $0.5$. The ControlNet model for the body outpainting stage is initialized from the official Openpose-pretrained checkpoint and tuned for 5 epochs ($6000$ iterations) on our filtered version of LAION-Human. Both models are trained on the Nvidia A1000 GPU with the batch size 32 and learning rate $1e-5$.

\subsection{Evaluation Metrics}

To evaluate the performance of the proposed approach we measure three aspects of the generation: pose accuracy including isolated evaluation of the hand poses, text-image consistency, and image quality. Pose accuracy is measured by Distance-based Average Precision (DAP) \cite{mscoco2014} and Mean Per Joint Position Error (MPJPE), calculated between the ground truth keypoints and the ones predicted with Mediapipe from the generated images. The core idea behind DAP is to mimic the evaluation metrics for object detection, namely Average Precision (AP) and Average Recall (AR). Originally, AP and AR use the Intersection over Union (IoU) measure for bounding boxes, thresholded at different levels, to match the ground truth and predicted objects. In the case of keypoints, IoU is replaced with a distance-based Object Keypoint Similarity (OKS) measure. In addition, MPJPE evaluates the average Euclidean distance between the predicted and ground truth joint positions.

\begin{figure}[t!]
    \centering
    \includegraphics[width=0.48\textwidth]{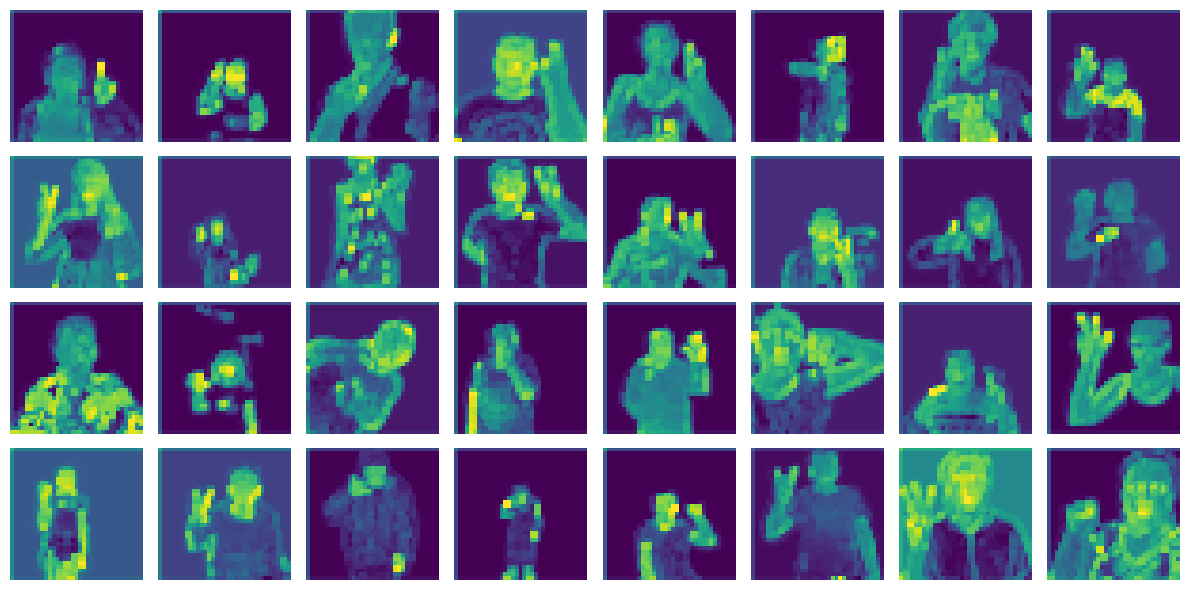}
    \caption{Visualization of the InceptionV3 convolutional features from the layer with feature dimension 192.}
    \label{fig:inception_features_192}
\end{figure}

Fréchet Inception Distance (FID \cite{gans2017fid}) and Kernel Inception Distance (KID \cite{Demystifying_mmd_gans2018kid}) are well established metrics that show the overall quality of the synthesis by comparing the distributions of Inception \cite{Szegedy2015inception} features extracted from ground-truth and generated images. The FID and KID can be calculated over the features from different layers of the Inception network with the choice of a layer impacting the sensitivity of the metrics to various aspects of image quality and diversity. As this work aims to improve hand generation in diffusion models, we are particularly interested in the quality of hand structure and patters associated with fingers. With this in mind, we explore the Inception features from different layers and identify the feature dimension 192 as the most suitable for our evaluation. We use the Torchmetrics \cite{Detlefsen2022torchmetrics} implementation of FID and KID and report the results for the chosen feature dimensionality. The features from a feature dimension 192 are visualized in Fig. \ref{fig:inception_features_192}.

Finally, we use the CLIP \cite{radford2021learning} similarity score (CLIPSIM) to measure the consistency between the input text prompt and generated images by projecting both into a shared latent space and calculating the distance between the embeddings.

\subsection{Results}

The proposed approach is compared to the recent state-of-the-art diffusion-based models, namely SD \cite{rombach2022high}, HandRefiner \cite{lu2023handrefiner}, HumanSD \cite{ju2023humansd}, T2I-Adapter \cite{mou2024t2i} and ControlNet \cite{zhang2023adding}. Following the HandRefiner evaluation setting, we randomly sample $12,000$ images from the \hagrid{} test set, keeping the original gesture distribution, and use them for the comparison. The quantitative results are summarized in Table \ref{tab:main_results}.

Firstly, we evaluate the precision of the produced poses for all the approaches by extracting the keypoints from the generated images with Mediapipe and comparing them to the ground truth. We report DAP and MPJPE across all 133 keypoints of the full body (17 for body, 68 for face, 21 for each hand, 6 for feet), as well as separately for 42 hand keypoints. The superiority of the proposed approach in pose controllability is demonstrated by a $30.5\%$ improvement from the baselines for the full body and $92.3\%$ improvement for the hand DAP. We also outperform the baselines in terms of MPJPE by $50\%$ for the full body and $40\%$ for the hand keypints. It is worth noting that SD and HandRefiner do not allow for pose conditioning and only base the generation on the text prompt. Text prompt conditioning, being an extremely weak guidance for the pose, results in $0.0$ DAP. This is because the predicted keypoints are too far from the corresponding ground-truth points for them to be associated with the same body parts by the algorithm.

\begin{figure}[b!]
    \centering
    \includegraphics[width=0.48\textwidth]{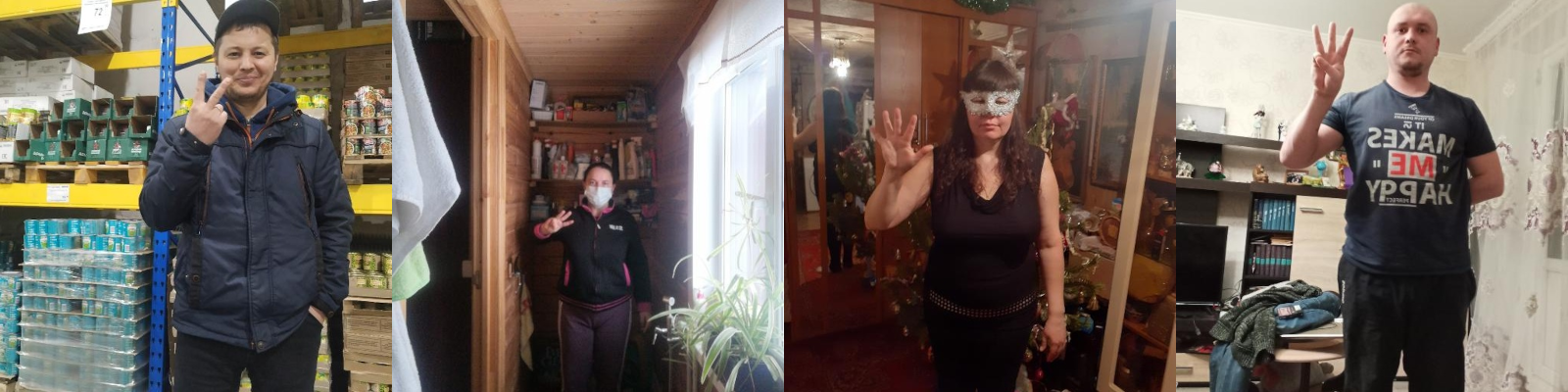}
    \caption{Examples of images from the \hagrid{} dataset with severe background clutter.}
    \label{fig:hagrid_bg_clutter}
\end{figure}

Initial experiments measuring the image quality showed poor performance despite excellent qualitative results. After investigation, it became apparent that samples from \hagrid{} often suffer from severe background clutter (see Fig. \ref{fig:hagrid_bg_clutter}). In the same way the Inception convolutional features, used for FID and KID computation, are sensitive to hand structure, they are also sensitive to clutter in the background. This sensitivity introduces noise to the evaluation metrics. With this in mind, we segment the background out using SAM to ensure a more fair and targeted evaluation of human generation quality. ``FID fg'' and ``KID fg'' in  Table \ref{tab:main_results} report the results on the images with background removed. In this human-centric setting, the proposed approach outperforms the baselines with a $22\%$ improvement. Similarly, our model shows higher results in text-to-image consistency.

\begin{table*}[t]
    \normalsize
    \centering
    \caption{Quantitative evaluation results on \hagrid{} dataset}
    \label{tab:main_results}
    \newcommand\highred[1]{\textcolor{red}{#1}}



\setlength\tabcolsep{3pt}%
\begin{tabular}{l||*{4}{c}|*{1}{c}|*{2}{c}}
    \toprule
    & \multicolumn{4}{c|}{\textbf{Pose Accuracy}} & \multicolumn{1}{c|}{} & \multicolumn{2}{c}{\textbf{Image Quality}}  \\
    \textbf{Method} & \textbf{DAP} $\uparrow$ & \textbf{DAP hands} $\uparrow$ & \textbf{MPJPE} $\downarrow$ & \textbf{MPJPE hands} $\downarrow$ & \textbf{CLIPSIM} $\uparrow$ & \textbf{FID fg} $\downarrow$ & \textbf{KID fg} $\downarrow$ \\
    \midrule
    Stable Diffusion \cite{rombach2022high}   & 0.00  & 0.00 & 0.381 & 0.469 & 32.94 & 2.40 &  1.28 $\pm$ 0.30 \\
    HandRefiner \cite{lu2023handrefiner}      & 0.00  & 0.00 & 0.380 & 0.466 & 32.95 & 2.33 & 1.17 $\pm$ 0.28 \\
    T2I-Adapter \cite{mou2024t2i}    & 0.06  & 0.11 & 0.179 & 0.216 & 33.09 & 2.39 & 1.29 $\pm$ 0.27 \\
    HumanSD \cite{ju2023humansd}              & 0.31  & 0.04 & 0.121 & 0.236 & 32.80 & 2.82 & 1.58 $\pm$ 0.29 \\
    ControlNet \cite{zhang2023adding}         & 0.59  & 0.39 & 0.094 & 0.135 & 32.86 & 2.34 & 1.46 $\pm$ 0.34 \\
    \textit{Ours}                             & $\mathbf{{0.77}}_{\highred{30.5\% \uparrow}}$ & $\mathbf{{0.75}}_{\highred{92.3\% \uparrow}}$ & $\mathbf{{0.047}}_{\highred{50\% \uparrow}}$ & $\mathbf{{0.081}}_{\highred{40\% \uparrow}}$ & \textbf{34.01} & \textbf{1.81} & $\mathbf{0.75 \pm 0.045}$ \\
    \bottomrule
\end{tabular}

\end{table*}

\subsection{Ablation Study}
\label{sec:ablation_study}

\begin{figure}[b!]
    \centering
    \includegraphics[width=0.48\textwidth]{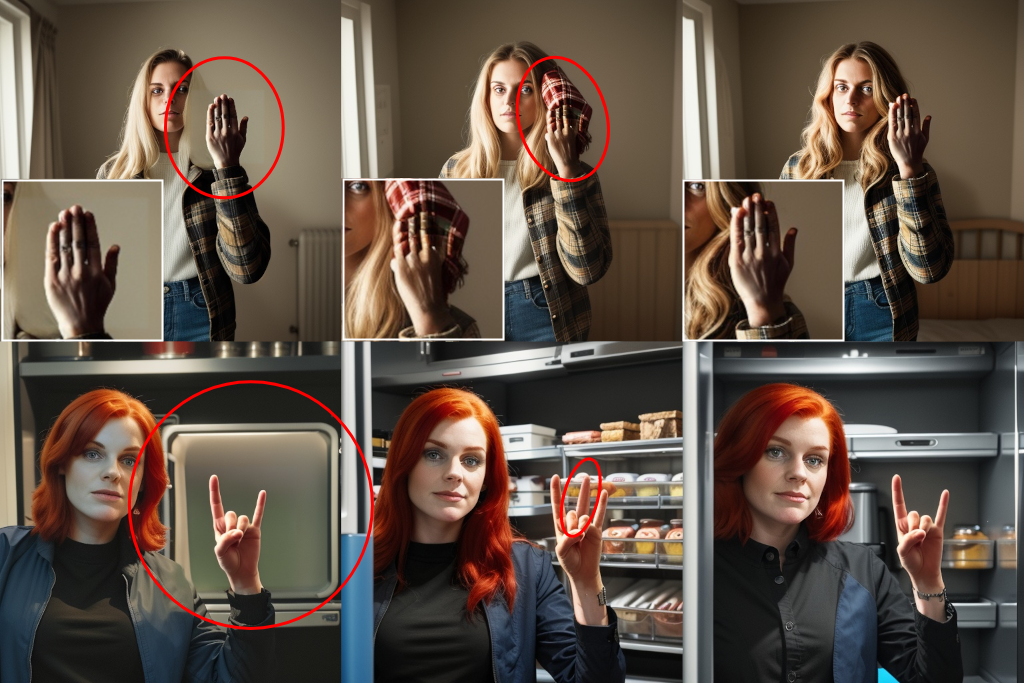}
    \caption{Qualitative comparison of three blending methods: bounding box (left), naive (middle) and sequential mask expansion (right).}
    \label{fig:blending-ablation-main}
\end{figure}

 It is crucial to employ a reliable blending strategy to combine the results of the hand generator and body outpainter in a harmonious and coherent way. To demonstrate the efficiency of the sequential mask expansion strategy, proposed in Section \ref{sec:blending}, we compare it to two alternative approaches: (1) \textit{bounding box blending} and (2) \textit{naive blending}. (1) defines the area outside the square hand region on the canvas as the outpainting region whereas (2) creates the outpainting mask by simply inverting the segmentation mask predicted by the hand generator. In all three cases, the last two steps of the diffusion process are performed with a full mask to smoothen the transitions between the regions. To compare the blending approaches, we randomly sample 500 images from the \hagrid{} test set, following the original gesture distribution, and measure FID, DAP and MPJPE between the generated images and the originals. It can be seen from Fig. \ref{fig:blending-ablation-main} that all three strategies are able to blend the hand and the body coherently. However, (1) does not allow to fully wash out the bounding box region and causes discoloration around the hands, corruption of the head and face, if hands are located in close proximity, and ``boxy'' background artifacts. At the same time, (2) tends to produce anomalies on the border of the hand region that include erroneous extensions of the hands, handheld objects and hallucinated textures. The proposed blending strategy allows us to preserve the area around the hand and eliminate artifacts on the border of the outpainted region. The numerical evaluation results in Table \ref{tab:blending_ablation} further demonstrate the sequential mask expansion mechanism outperforming the alternatives in terms of both quality and pose precision metrics. Please refer to the supplementary material for additional qualitative comparison of the three blending methods.

\begin{table}[t!]
    \normalsize
    \centering
    \caption{Ablation on the blending strategy.}
    \label{tab:blending_ablation}
    \setlength\tabcolsep{4pt}%
\begin{tabular}{l||*{3}{c}}
    \toprule
    \textbf{Method} & \textbf{FID} $\downarrow$ & \textbf{DAP} $\uparrow$ & \textbf{MPJPE} $\downarrow$  \\
    \midrule

    Bounding Box blending                               & 16.46 & 0.49 & 0.087    \\
    Naive blending                                      & 13.03 & 0.58 & 0.062     \\
    \textit{Sequential Mask Expansion}                  & \textbf{12.13} & \textbf{0.59} & \textbf{0.057}     \\
    
    \bottomrule
\end{tabular}
\end{table}


\section{LIMITATIONS AND CONCLUSIONS}

In this work we presented a novel approach to human image generation that addresses the issue of low-quality hand synthesis and lack of control over the resulting hand pose. The experimental evaluations on HaGRID dataset showed the increased performance of our approach in terms of both pose precision and image quality, comparing to a number of state-of-the-art diffusion-based approaches to image generation.

Although the proposed model produces impressive visual results, there are some limitations to it. We rely on a connectivity between arms and wrists in the input body keypoints. In cases where the hand keypoints are present but the arm is missing from the skeleton, the model may produce discontinuities in the generated image. This is due to the fact that the hand will be generated in the first stage of the process but the arm may not be present to connect to it.

Furthermore, the presented approach concentrates on cases where hands occupy a substantial area in the frame. This is because the spatial dimensions $64 \times 64$ of the SD latent space may be insufficient to accommodate the fine details of small hand masks during the outpainting step. Therefore, in the proposed setting, the quality of small hand regions may decrease. 

Currently, the results of the hand generator are decoded to the pixel space to be further encoded into latents again for the outpainting stage. As the VAE latent encoding-decoding procedure is lossy, it may result in a decreased quality of the hand region. It is also less efficient from the inference time standpoint. We leave bringing both stages of the process to a shared latent space for future work.

\section{ACKNOWLEDGEMENTS}

This work was supported by the SNSF project ‘SMILE II’ (CRSII5 193686), European Union’s Horizon2020 programme (‘EASIER’ grant agreement 101016982) and the Innosuisse IICT Flagship (PFFS-21-47). This work reflects only the authors view and the Commission is not responsible for any use that may be made of the information it contains.


\pagebreak

{\bibliographystyle{ieee}
\bibliography{egbib}}

\newpage



\overrideIEEEmargins



\title{\LARGE \bf
Giving a Hand to Diffusion Models: a Two-Stage Approach to Improving Conditional Human Image Generation \\
Supplementary Material
\vspace{-20pt}
}
\author{}

\maketitle
\thispagestyle{plain}

\subsection{Text Prompts}
\label{sec:sup_text_prompts}

All text prompts presented here have been automatically extracted from the HaGRID dataset test images using the LLaVA-v1.5-7b captioning model. The images in Fig. 1 of the main paper were generated using the following text prompts:
\begin{enumerate}
    \item \small{``The person in the image is a woman with dark hair, wearing a striped shirt. She is making a peace sign with her hands, and there is a pink wall behind her. The overall visual style of the image is casual and candid, capturing a moment of the woman's life.''}
    \item  \small{``The person in the image is a woman with short hair, wearing a gray shirt and black pants. She is holding up four fingers, possibly to make a statement or express her feelings. The overall visual style of the image is a close-up of the woman, focusing on her facial expression and hand gesture.''}
    \item  \small{``The image features a man with a beard and a blue shirt. He is sitting down and holding a cell phone up to his ear. The man appears to be making a funny face, possibly for the camera. The overall visual style of the image is casual and candid, capturing a moment of the man's life.''}
\end{enumerate}

The additional qualitative examples in Fig.\ref{fig:sup_more_qualitative_examples} were generated using the following text prompts:
\begin{enumerate}
    \item  \small{``The person in the image is an older man wearing glasses and a blue shirt. He is making a peace sign with his hand. The overall visual style of the image is a close-up of the man, focusing on his facial features and hand gesture.''}
    \item  \small{``The image features a young woman wearing a white shirt with a cartoon character, specifically a tiger, on it. She is making a "peace" sign with her hand. The overall visual style of the image is casual and informal, with the focus on the woman and her gesture.''}
    \item  \small{``The image features a woman with short hair, wearing a gray shirt and a black sweater. She is holding a cell phone to her ear, possibly engaged in a conversation. The overall visual style of the image is simple and straightforward, focusing on the woman and her activity.''}
    \item  \small{``The person in the image is a woman wearing a green dress, standing in an office setting. She is making a hand gesture, possibly a peace sign or a rock and roll hand gesture, while looking at the camera. The overall visual style of the image is a close-up of the woman, emphasizing her appearance and the hand gesture she is making.''}
    \item  \small{``The person in the image is a man wearing a gray shirt and standing in a room. He is making a hand gesture, possibly a rock sign or a peace sign, while posing for the picture. The room appears to be a living space, with a couch visible in the background. The overall visual style of the image is casual and candid, capturing a moment of the man's expression.''}
    \item  \small{``The person in the image is a man wearing a blue sweater and glasses. He is making a peace sign with his hand. The overall visual style of the image is a close-up of the man, focusing on his facial expression and hand gesture. The setting appears to be indoors, possibly in a hallway or a room.''}
    \item  \small{``The image features a young woman with short black hair, wearing a purple shirt. She is making a thumbs-down gesture, possibly expressing her disapproval or disagreement with something. The woman is standing in a room, and there is a potted plant in the background, adding a touch of greenery to the scene. The overall visual style of the image is casual and candid, capturing the woman's spontaneous reaction to a situation.''}
    \item  \small{``The image features a woman with long, curly hair, wearing a red shirt. She is making a ""thumbs up"" gesture with her hand, which is a common way to express approval, agreement, or excitement. The overall visual style of the image is casual and candid, as it appears to be a selfie taken by the woman herself.''}
\end{enumerate}

\vspace{5pt}
\subsection{Additional Qualitative Results}
\label{sec:sup_more_figures}

\begin{figure}[h]
    \centering
    \includegraphics[width=0.45\textwidth]{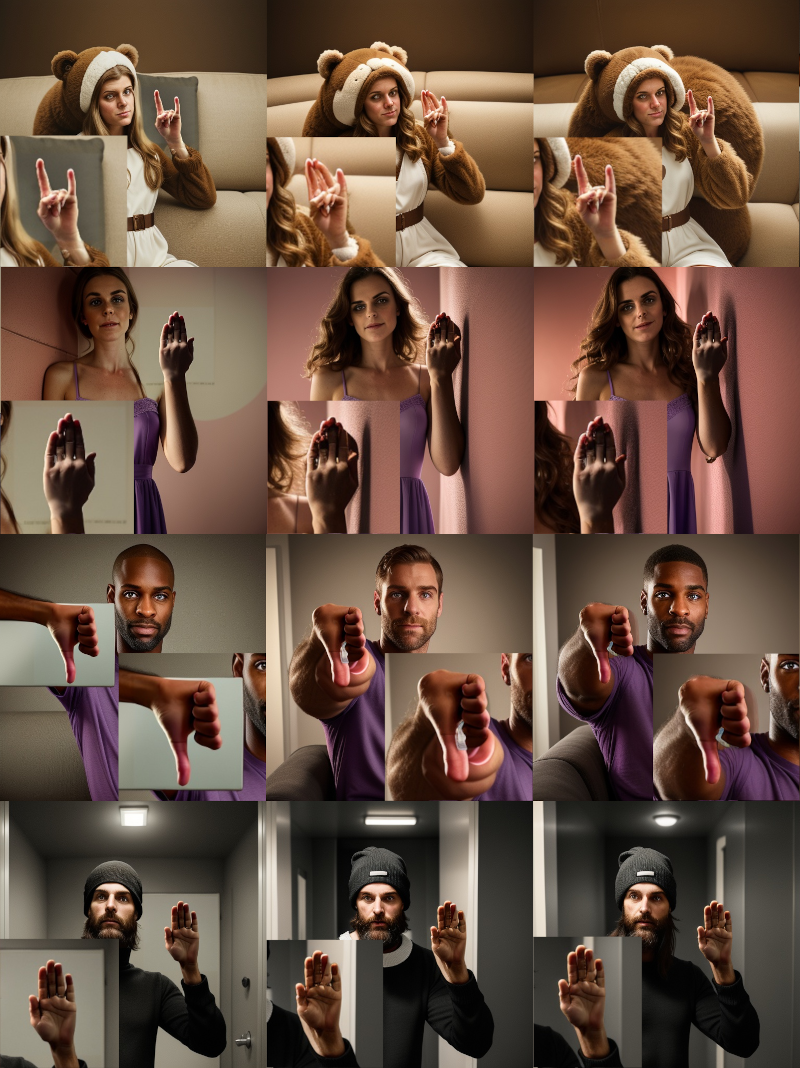}
    \caption{Additional qualitative comparison of the three blending methods: bounding box (left), naive (middle) and sequential mask expansion (right).}
    \label{fig:blending_ablation_more_examples}
\end{figure}

\begin{figure*}[t!]
    \centering
    \setlength{\tabcolsep}{0pt} 
    \begin{tabular}{ccccccc}
        \includegraphics[width=0.14\linewidth, height=0.14\linewidth]{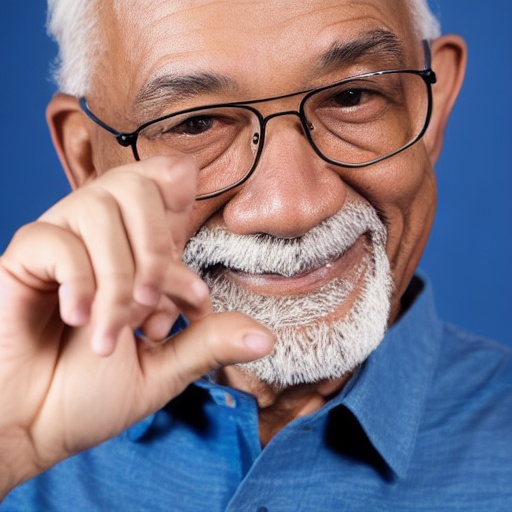} &
        \includegraphics[width=0.14\linewidth, height=0.14\linewidth]{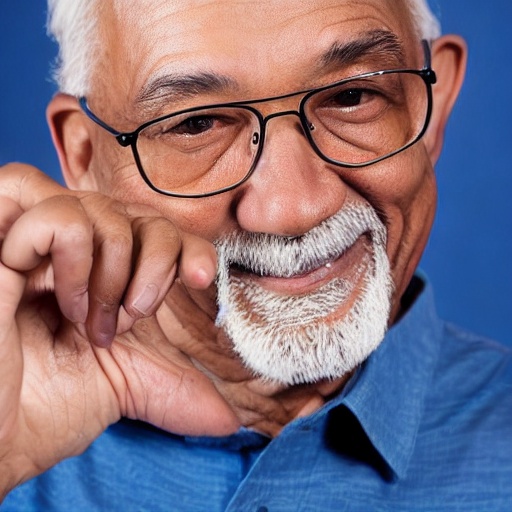} &
        \includegraphics[width=0.14\linewidth, height=0.14\linewidth]{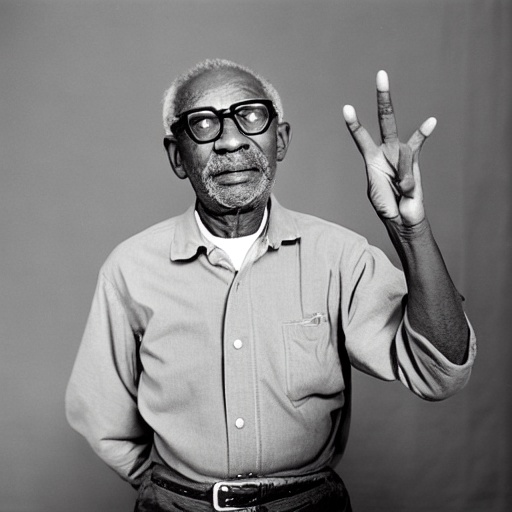} &
        \includegraphics[width=0.14\linewidth, height=0.14\linewidth]{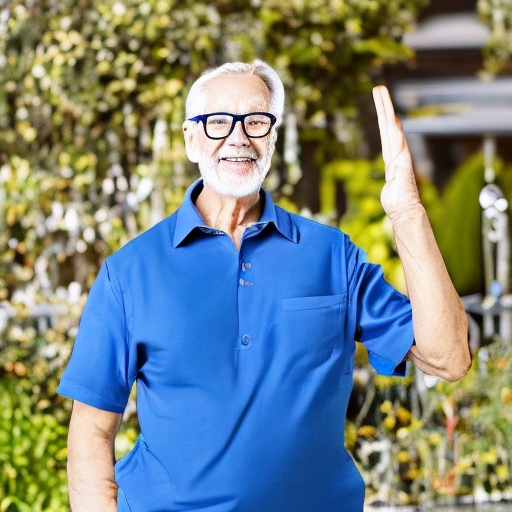} &
        \includegraphics[width=0.14\linewidth, height=0.14\linewidth]{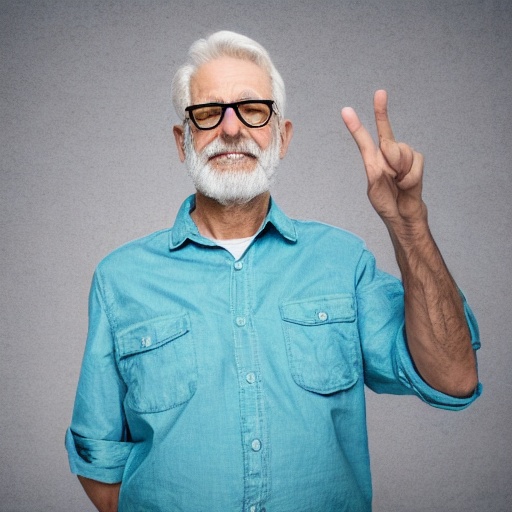} &
        \includegraphics[width=0.14\linewidth, height=0.14\linewidth]{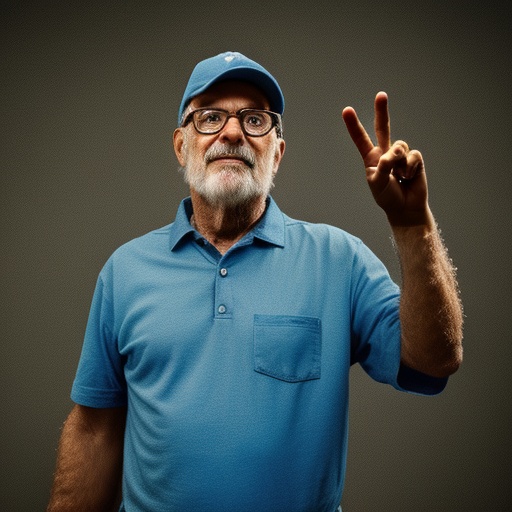} &
        \includegraphics[width=0.14\linewidth, height=0.14\linewidth]{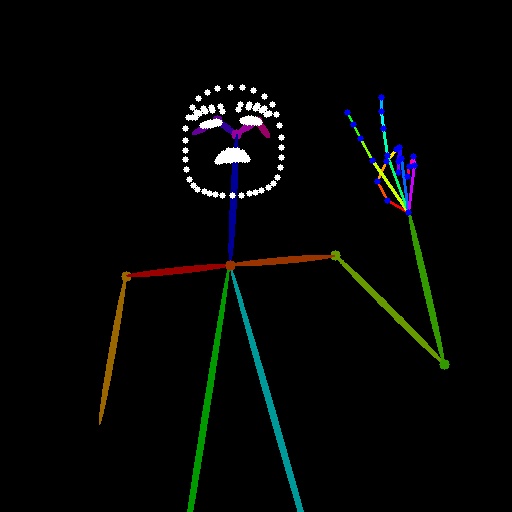} \\

        \includegraphics[width=0.14\linewidth, height=0.14\linewidth]{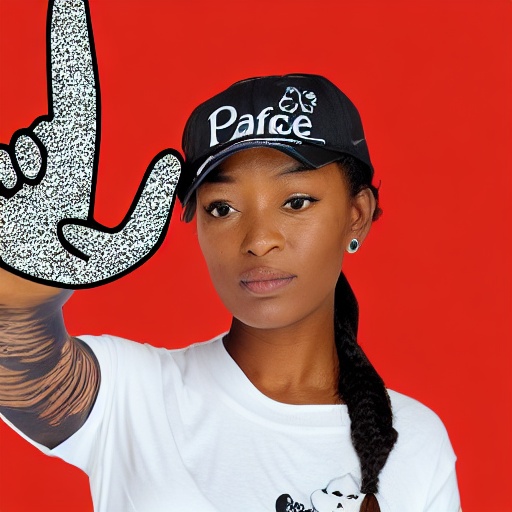} &
        \includegraphics[width=0.14\linewidth, height=0.14\linewidth]{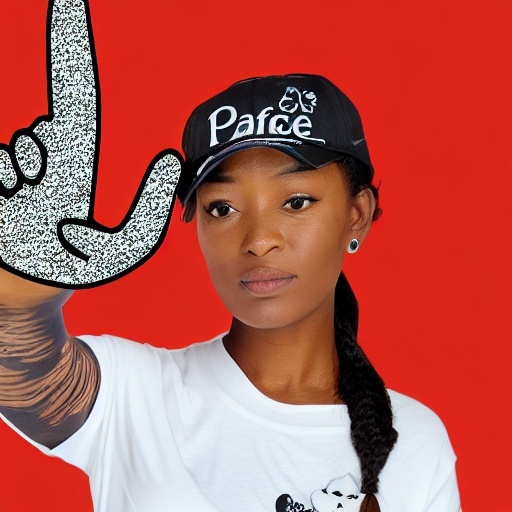} &
        \includegraphics[width=0.14\linewidth, height=0.14\linewidth]{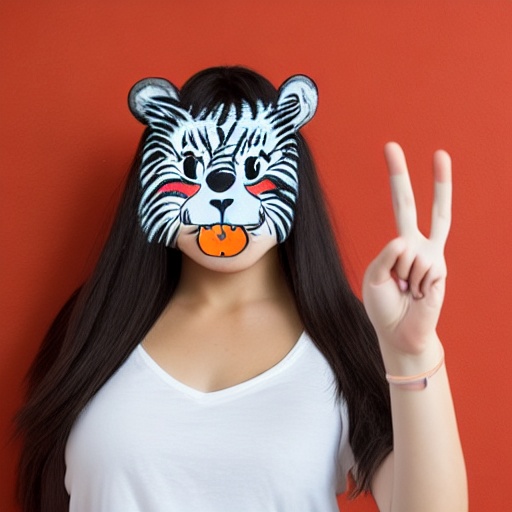} &
        \includegraphics[width=0.14\linewidth, height=0.14\linewidth]{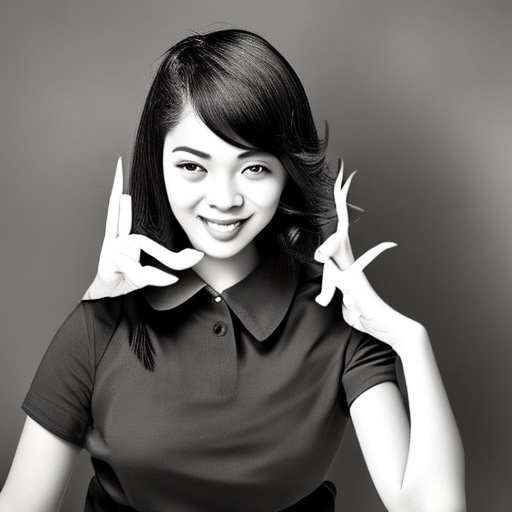} &
        \includegraphics[width=0.14\linewidth, height=0.14\linewidth]{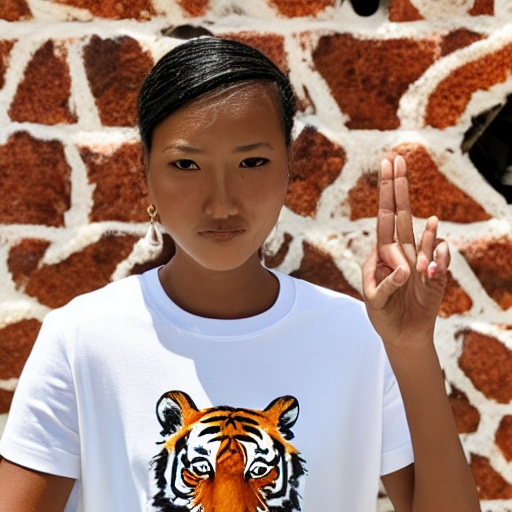} &
        \includegraphics[width=0.14\linewidth, height=0.14\linewidth]{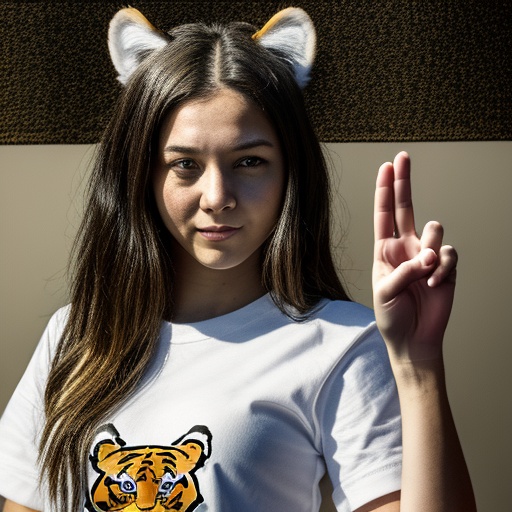} &
        \includegraphics[width=0.14\linewidth, height=0.14\linewidth]{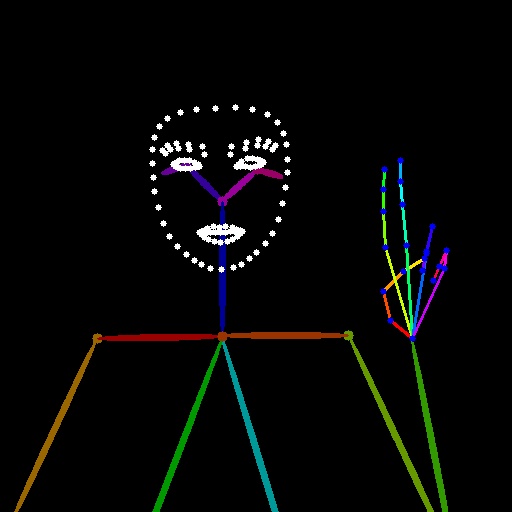} \\

        \includegraphics[width=0.14\linewidth, height=0.14\linewidth]{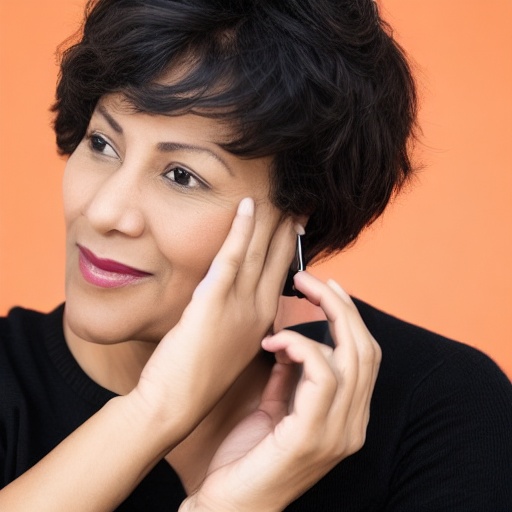} &
        \includegraphics[width=0.14\linewidth, height=0.14\linewidth]{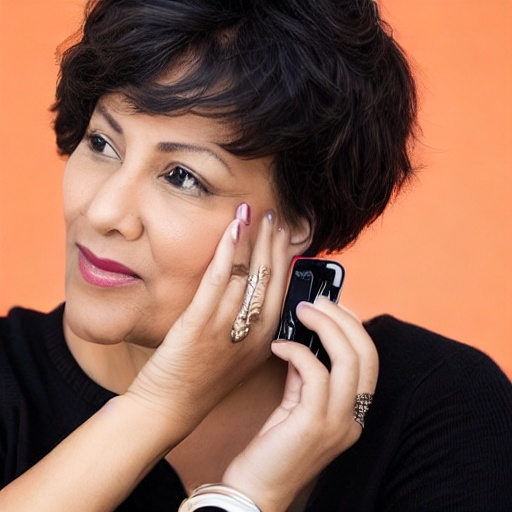} &
        \includegraphics[width=0.14\linewidth, height=0.14\linewidth]{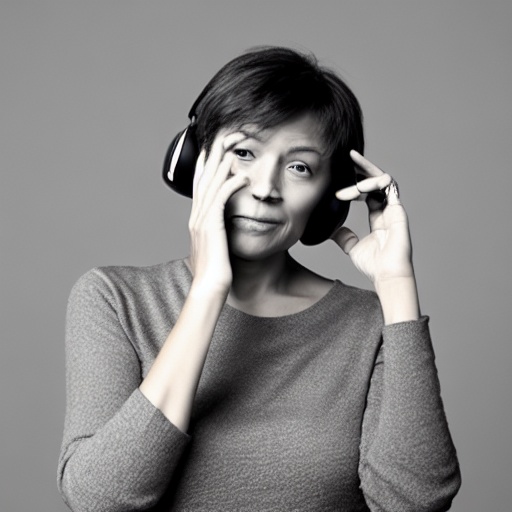} &
        \includegraphics[width=0.14\linewidth, height=0.14\linewidth]{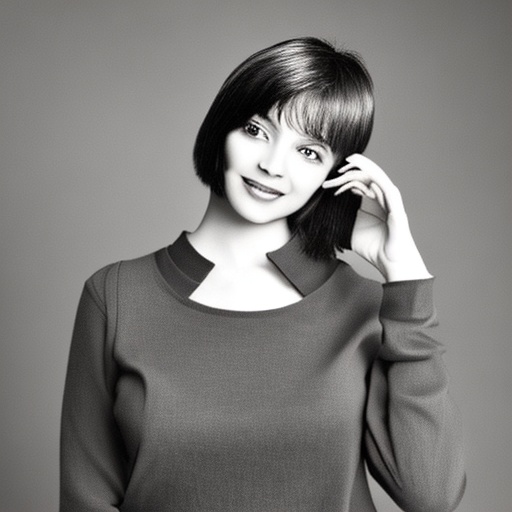} &
        \includegraphics[width=0.14\linewidth, height=0.14\linewidth]{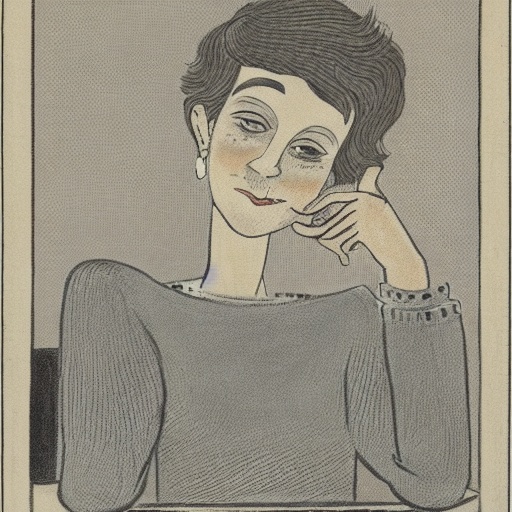} &
        \includegraphics[width=0.14\linewidth, height=0.14\linewidth]{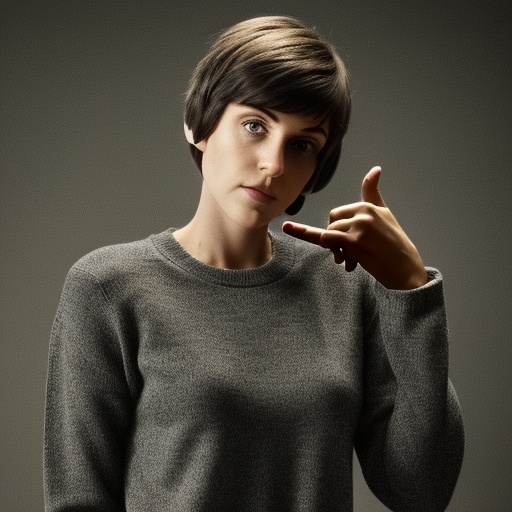} &
        \includegraphics[width=0.14\linewidth, height=0.14\linewidth]{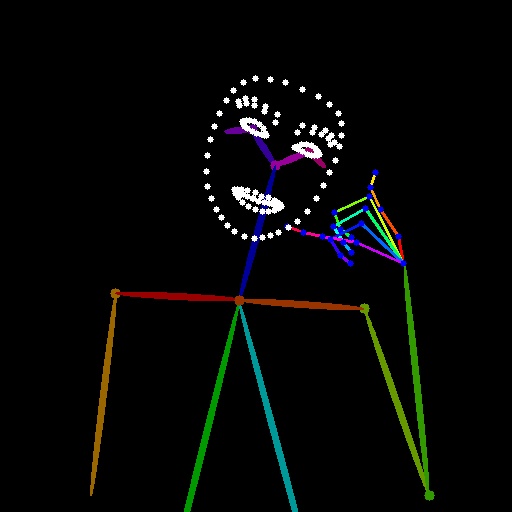} \\
        
        \includegraphics[width=0.14\linewidth, height=0.14\linewidth]{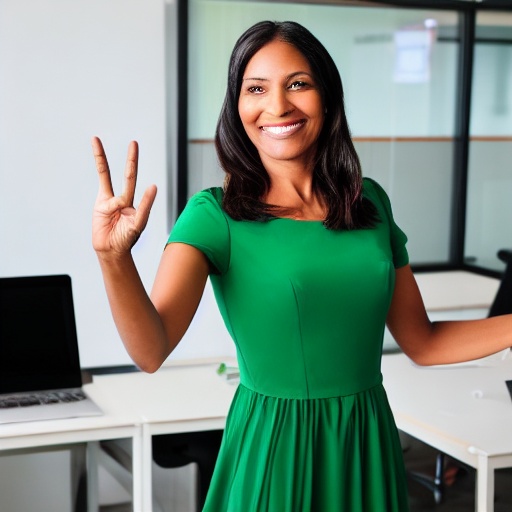} &
        \includegraphics[width=0.14\linewidth, height=0.14\linewidth]{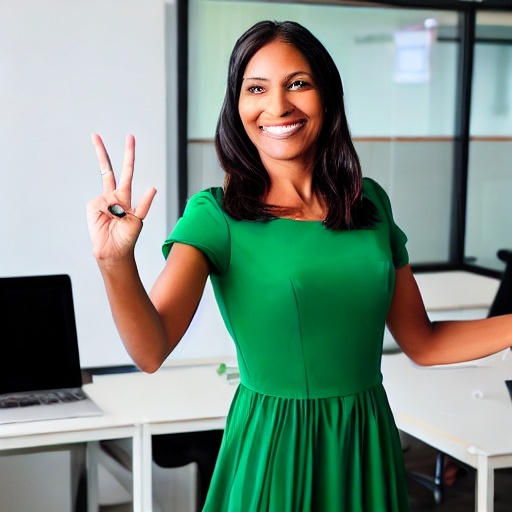} &
        \includegraphics[width=0.14\linewidth, height=0.14\linewidth]{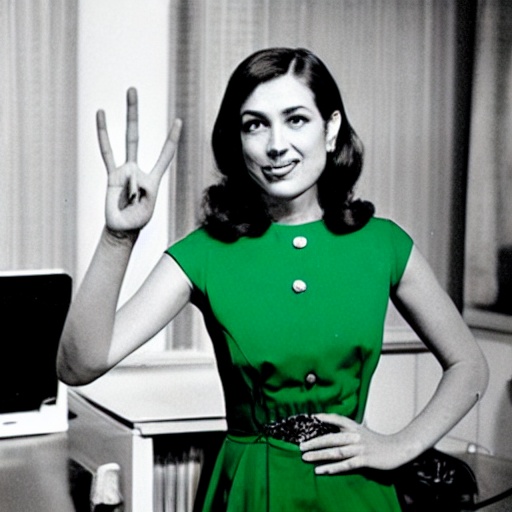} &
        \includegraphics[width=0.14\linewidth, height=0.14\linewidth]{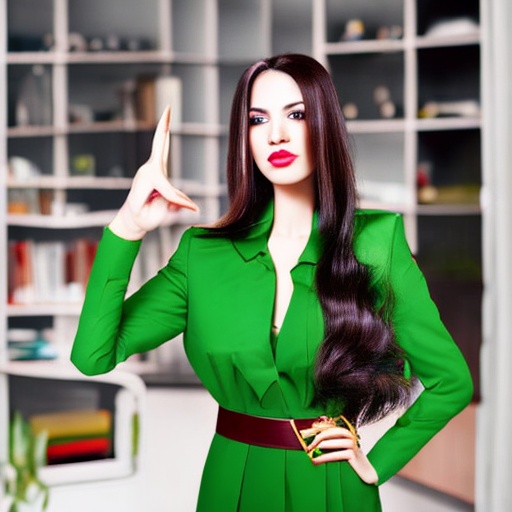} &
        \includegraphics[width=0.14\linewidth, height=0.14\linewidth]{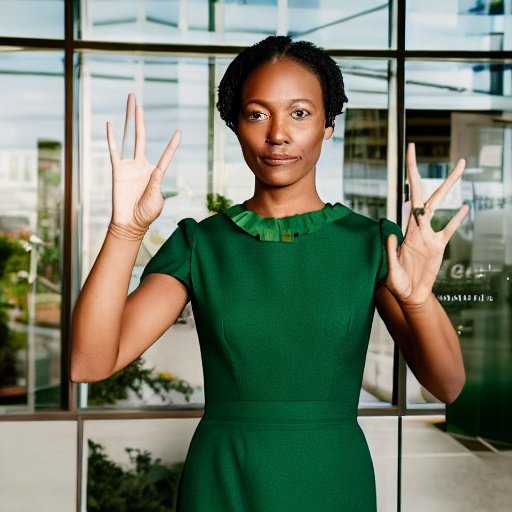} &
        \includegraphics[width=0.14\linewidth, height=0.14\linewidth]{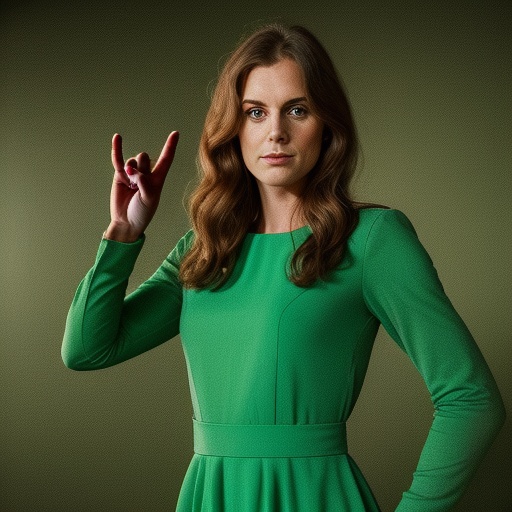} &
        \includegraphics[width=0.14\linewidth, height=0.14\linewidth]{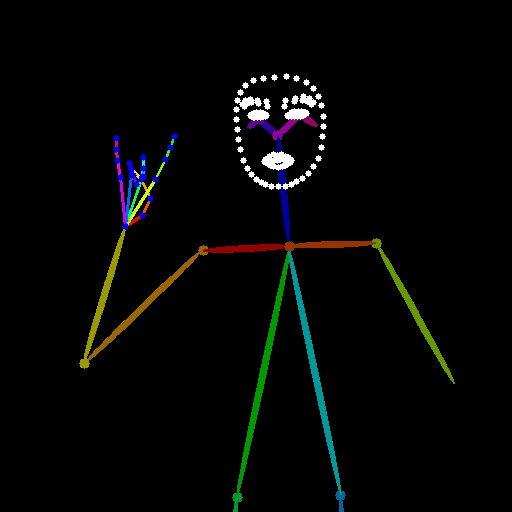} \\
        
        \includegraphics[width=0.14\linewidth, height=0.14\linewidth]{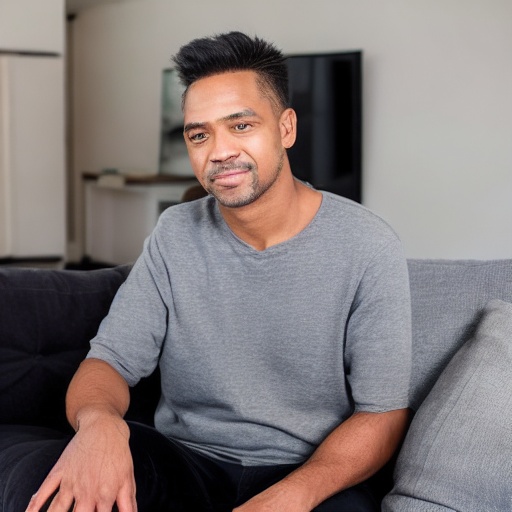} &
        \includegraphics[width=0.14\linewidth, height=0.14\linewidth]{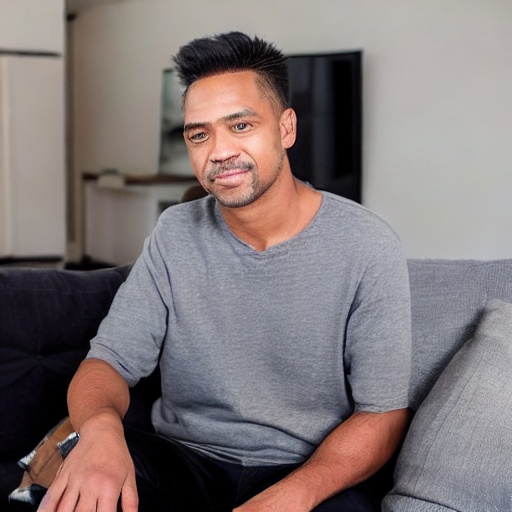} &
        \includegraphics[width=0.14\linewidth, height=0.14\linewidth]{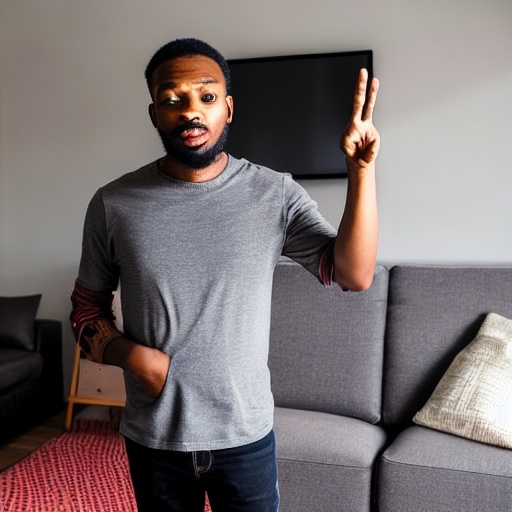} &
        \includegraphics[width=0.14\linewidth, height=0.14\linewidth]{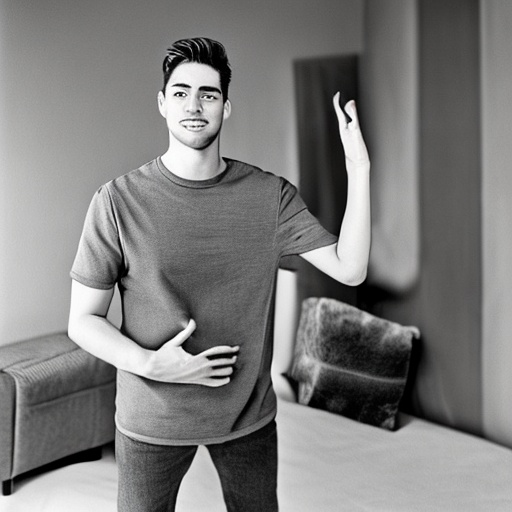} &
        \includegraphics[width=0.14\linewidth, height=0.14\linewidth]{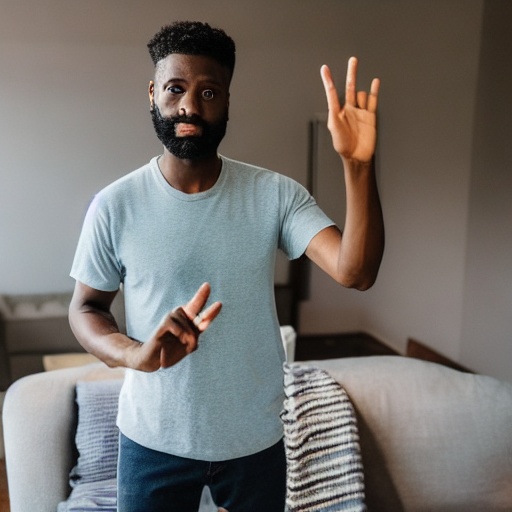} &
        \includegraphics[width=0.14\linewidth, height=0.14\linewidth]{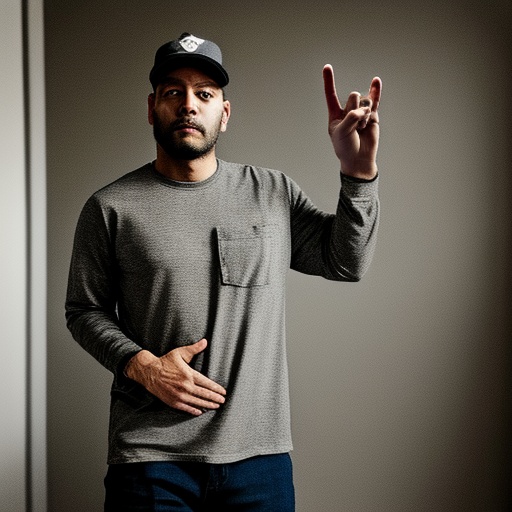} &
        \includegraphics[width=0.14\linewidth, height=0.14\linewidth]{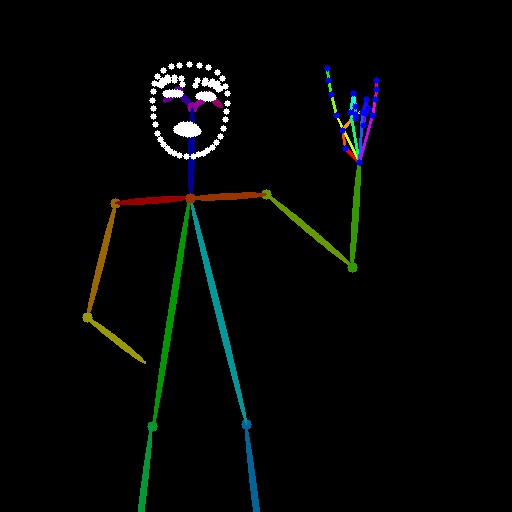} \\

        \includegraphics[width=0.14\linewidth, height=0.14\linewidth]{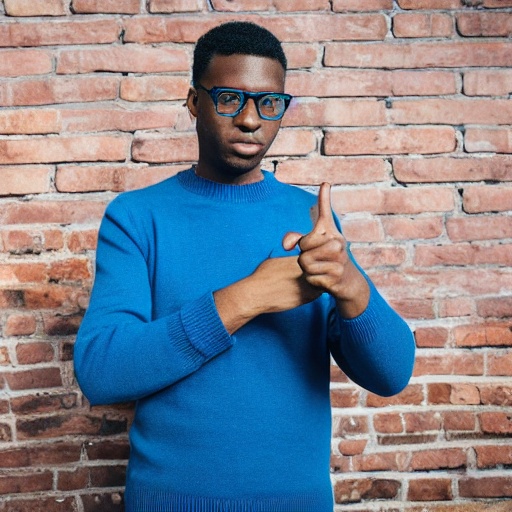} &
        \includegraphics[width=0.14\linewidth, height=0.14\linewidth]{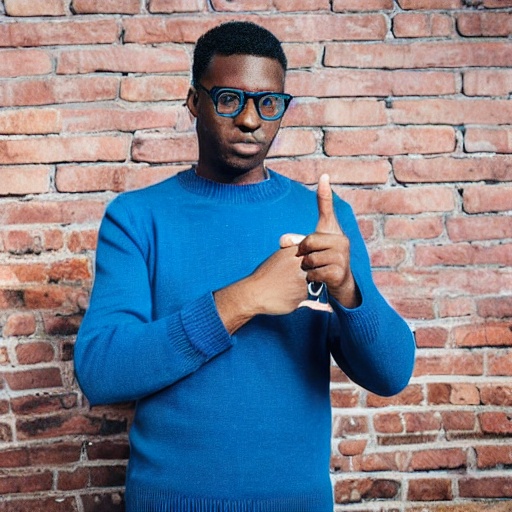} &
        \includegraphics[width=0.14\linewidth, height=0.14\linewidth]{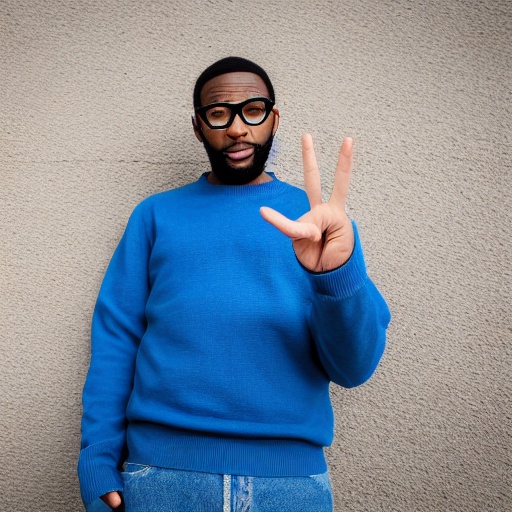} &
        \includegraphics[width=0.14\linewidth, height=0.14\linewidth]{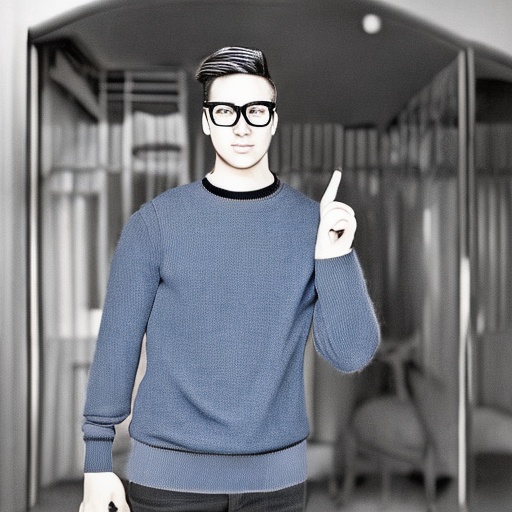} &
        \includegraphics[width=0.14\linewidth, height=0.14\linewidth]{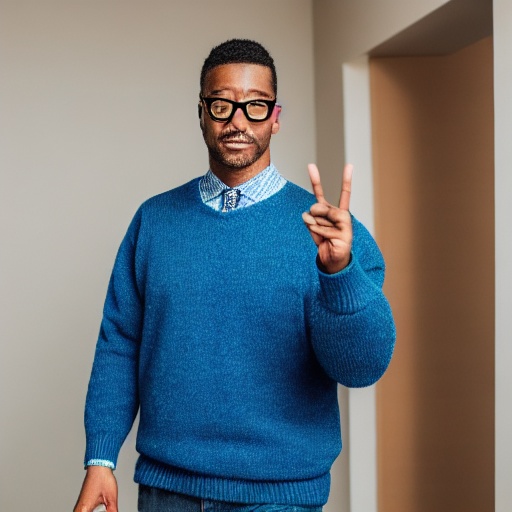} &
        \includegraphics[width=0.14\linewidth, height=0.14\linewidth]{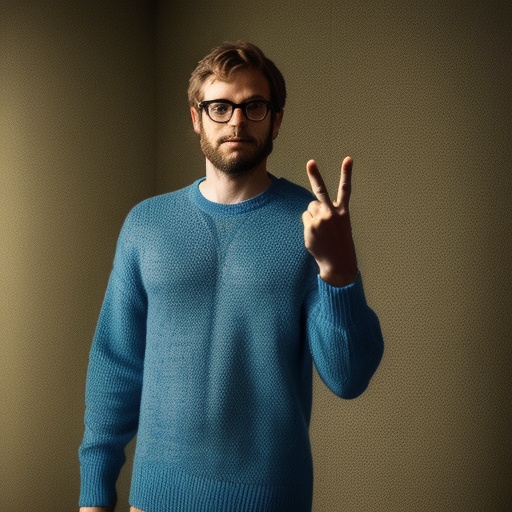} &
        \includegraphics[width=0.14\linewidth, height=0.14\linewidth]{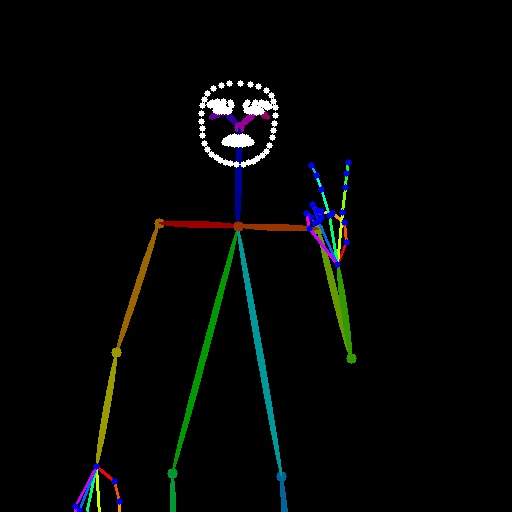} \\
        
        \includegraphics[width=0.14\linewidth, height=0.14\linewidth]{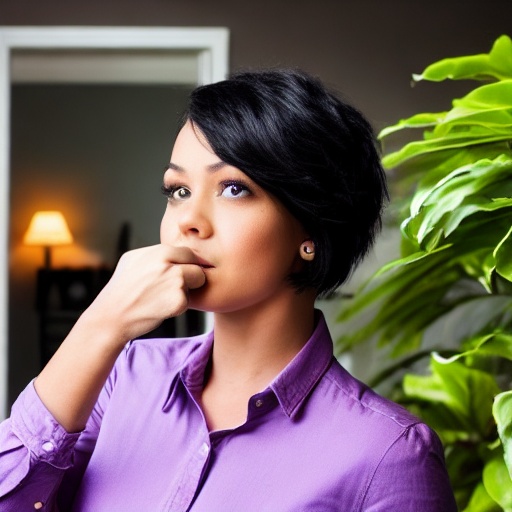} &
        \includegraphics[width=0.14\linewidth, height=0.14\linewidth]{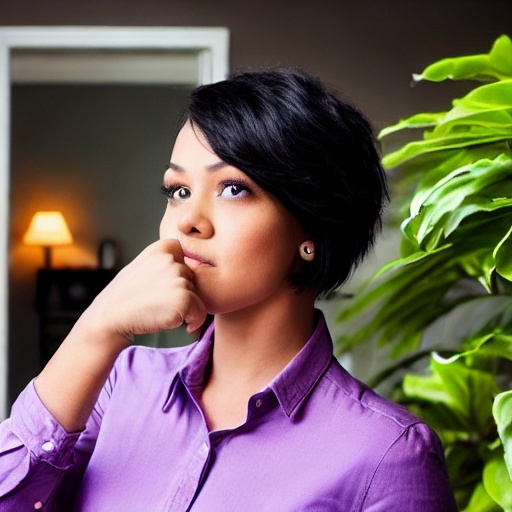} &
        \includegraphics[width=0.14\linewidth, height=0.14\linewidth]{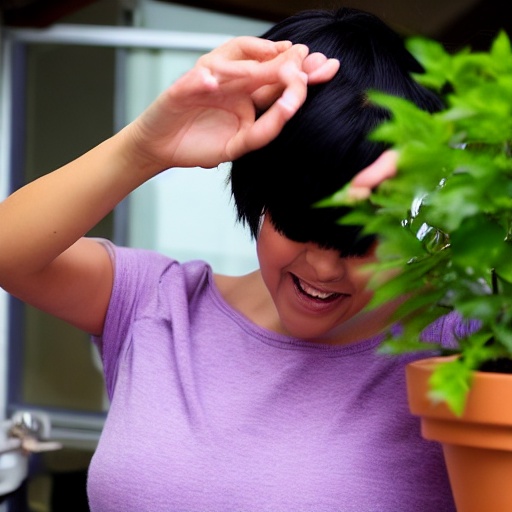} &
        \includegraphics[width=0.14\linewidth, height=0.14\linewidth]{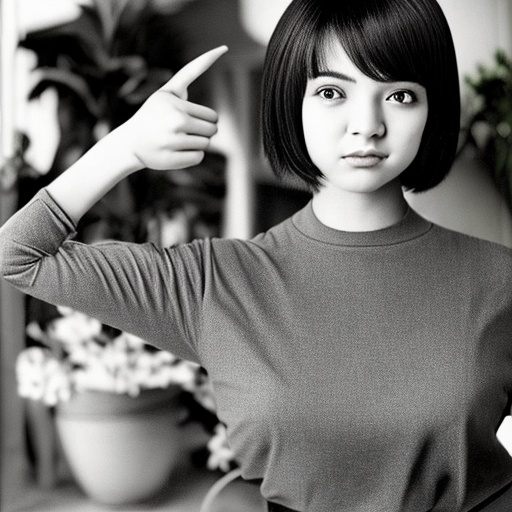} &
        \includegraphics[width=0.14\linewidth, height=0.14\linewidth]{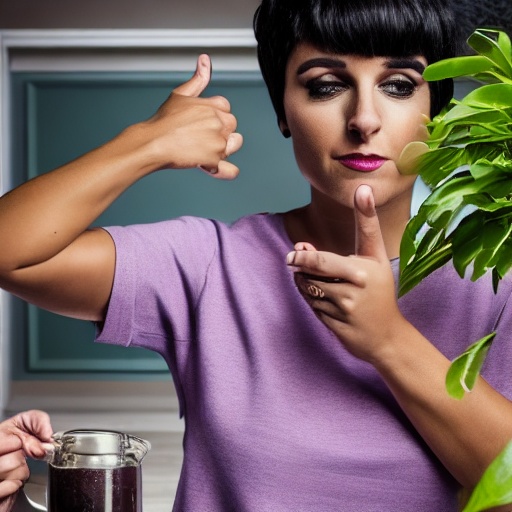} &
        \includegraphics[width=0.14\linewidth, height=0.14\linewidth]{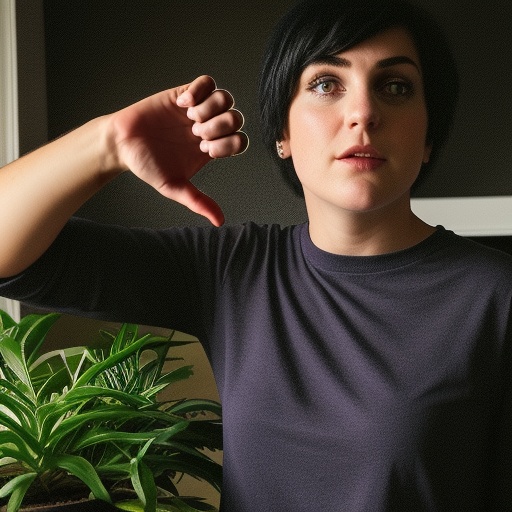} &
        \includegraphics[width=0.14\linewidth, height=0.14\linewidth]{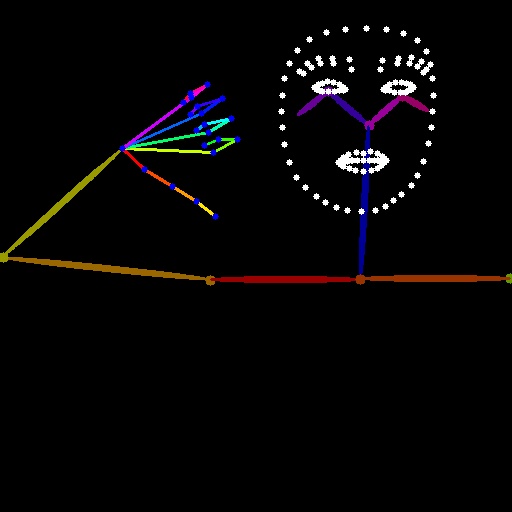} \\

        \includegraphics[width=0.14\linewidth, height=0.14\linewidth]{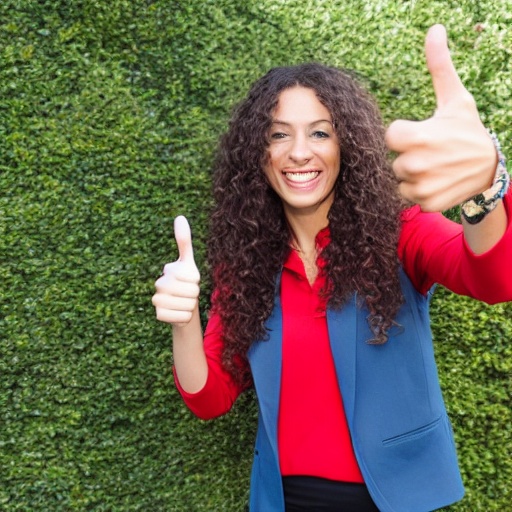} &
        \includegraphics[width=0.14\linewidth, height=0.14\linewidth]{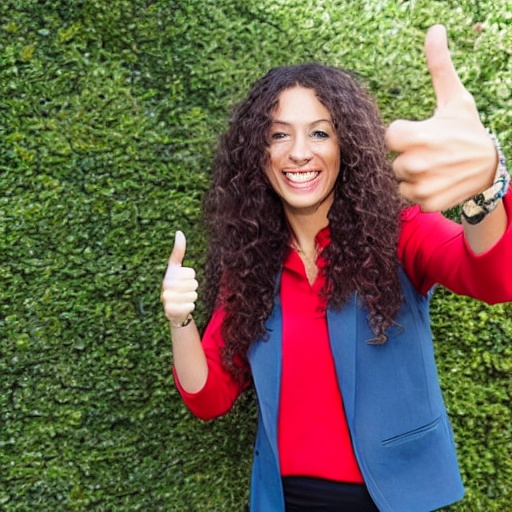} &
        \includegraphics[width=0.14\linewidth, height=0.14\linewidth]{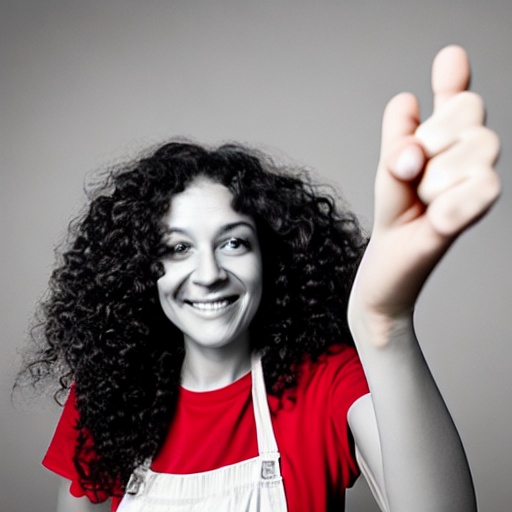} &
        \includegraphics[width=0.14\linewidth, height=0.14\linewidth]{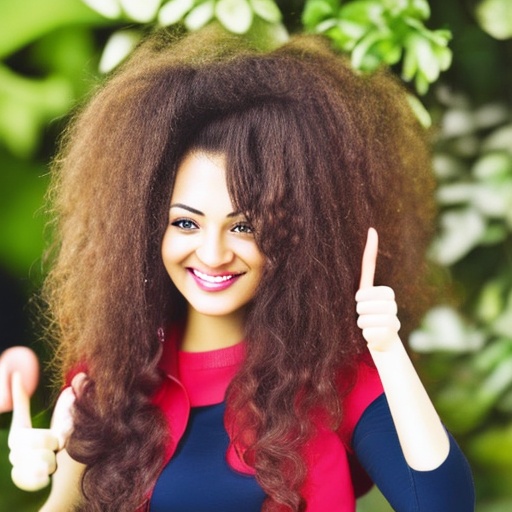} &
        \includegraphics[width=0.14\linewidth, height=0.14\linewidth]{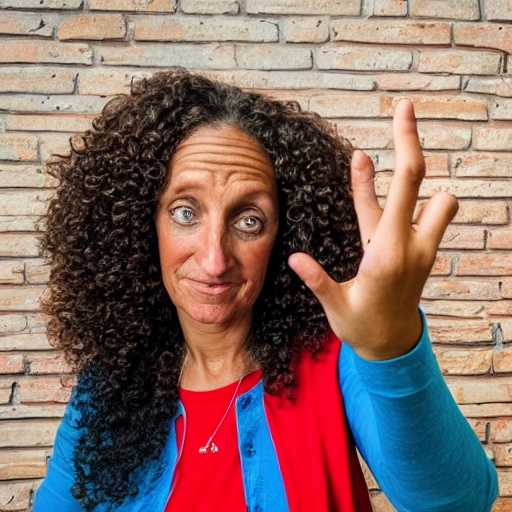} &
        \includegraphics[width=0.14\linewidth, height=0.14\linewidth]{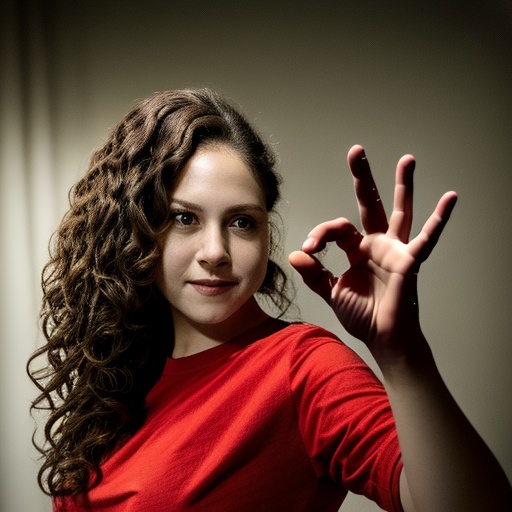} &
        \includegraphics[width=0.14\linewidth, height=0.14\linewidth]{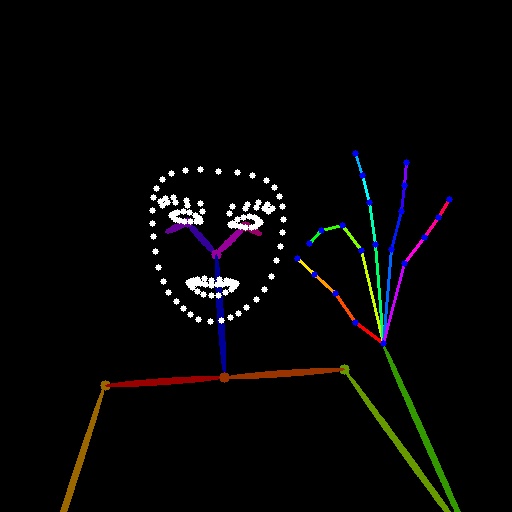} \\
        
        {\small Stable Diffusion}  & {\small HandRefiner}  & {\small T2I-Adapter}  & {\small HumanSD} & {\small ControlNet} & {\small \textit{Ours}} & {\small Pose condition} \\
    \end{tabular}

    \caption{\small Additional examples of images, generated by the proposed method (column 6) and the state-of-the-art diffusion models (columns 1 to 5), given the pose condition (final column) and the text description.}
    \label{fig:sup_more_qualitative_examples}
\end{figure*}


\end{document}